\documentclass[sigconf]{acmart}

\AtBeginDocument{%
  \providecommand\BibTeX{{%
    \normalfont B\kern-0.5em{\scshape i\kern-0.25em b}\kern-0.8em\TeX}}}

\setcopyright{acmcopyright}
\copyrightyear{2020}
\acmYear{2020}
\acmDOI{10.1145/1122445.1122456}

\acmConference[MM '20] {28th ACM International Conference on Multimedia}{October 12--16, 2020}{Seattle, WA, USA.}
\acmBooktitle{28th ACM International Conference on Multimedia (MM '20), October 12--16, 2020, Seattle, WA, USA.}
\acmPrice{15.00}
\acmDOI{10.1145/3394171.3413624}
\acmISBN{978-1-4503-7988-5/20/10}

\acmSubmissionID{mmfp1308}

\usepackage{color}

\usepackage{subcaption}
\usepackage{graphicx}
\usepackage{multirow}
\usepackage{balance}

\begin{document}
\fancyhead{}

\title{Crossing You in Style:\\ Cross-modal Style Transfer from Music to Visual Arts}

% \affiliation{%

\author{Cheng-Che Lee}
\authornote{First two authors contributed equally to this research.}
\email
{nctusunnerli.cs06g@nctu.edu.tw}
\affiliation{
  \institution{National Chiao Tung University}
  \city{Hsinchu}
  \state{Taiwan}
}

\author{Wan-Yi Lin}
\email{softcat477@gmail.com}
\affiliation{%
  \institution{National Tsing Hua University}
  \city{Hsinchu}
  \state{Taiwan}
}

\author{Yen-Ting Shih}
\email{steven88sky@gapp.nthu.edu.tw}
\affiliation{%
  \institution{National Tsing Hua University}
  \city{Hsinchu}
  \state{Taiwan}
}

\author{Pei-Yi Patricia Kuo}
\email{pykuo@iss.nthu.edu.tw}
\affiliation{%
  \institution{National Tsing Hua University}
  \city{Hsinchu}
  \state{Taiwan}
}

\author{Li Su}
\email{lisu@iis.sinica.edu.tw}
\affiliation{%
  \institution{Academia Sinica}
  \city{Taipei}
  \state{Taiwan}
}

\renewcommand{\shortauthors}{Lee and Lin, et al.}

\begin{abstract}
  Music-to-visual style transfer is a challenging yet important cross-modal learning problem in the practice of creativity. Its major difference from the traditional image style transfer problem is that the style information is provided by music rather than images. Assuming that musical features can be properly mapped to visual contents through semantic links between the two domains, we solve the music-to-visual style transfer problem in two steps: music visualization and style transfer. The music visualization network utilizes an encoder-generator architecture with a conditional generative adversarial network to generate image-based music representations from music data. This network is integrated with an image style transfer method to accomplish the style transfer process. Experiments are conducted on WikiArt-IMSLP, a newly compiled dataset including Western music recordings and paintings listed by decades. By utilizing such a label to learn the semantic connection between paintings and music, we demonstrate that the proposed framework can generate diverse image style representations from a music piece, and these representations can unveil certain art forms of the same era. Subjective testing results also emphasize the role of the era label in improving the perceptual quality on the compatibility between music and visual content.
\end{abstract}



% \begin{figure}
%     \centering
%     \includegraphics[width=\linewidth]{Figure/teaser.pdf}
%     \caption{Caption}
%     \label{fig:my_label}
% \end{figure}

%     \centering
%     \includegraphics[width=0.99\linewidth]{Figure/teaser6.png}
%     \caption{Let music change the visual style of an image. For example, a spectrogram of Claude Debussy's music \emph{Sarabande} in \emph{Pour le piano, L. 95} (1901) transfers Alfred Jacob Miller's painting \emph{The Lake Her Lone Bosom Expands to the Sky} (1850) into an Impressionism-like color scheme through a neural network linking the semantic space shared by music and image.}
%     \label{fig:teaser}
%     \vspace{0.2cm}

\begin{teaserfigure}
    \centering
    \includegraphics[width=0.99\linewidth]{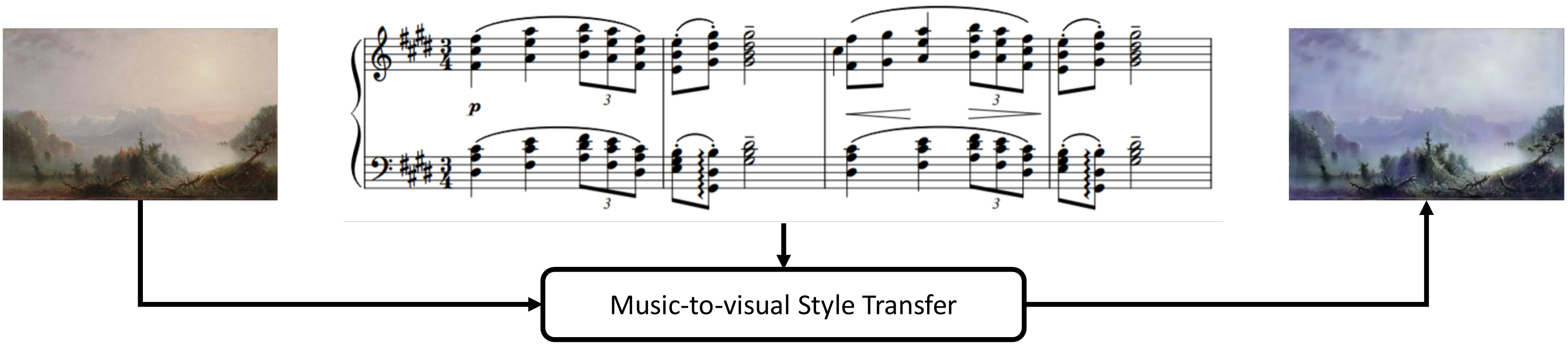}
    \caption{Let music change the visual style of an image. For example, a spectrogram of Claude Debussy's music \emph{Sarabande} in \emph{Pour le piano, L. 95} (1901) transfers Alfred Jacob Miller's painting \emph{The Lake Her Lone Bosom Expands to the Sky} (1850) into an Impressionism-like color scheme through a neural network linking the semantic space shared by music and image.}
    \label{fig:teaser}
\end{teaserfigure}

\maketitle

\section{Introduction}

Since Gatys \emph{et al.} proposed the neural algorithm for image style transfer \cite{gatys2016image}, deep learning-based style transfer has been extensively studied. Various types of neural network models now can modify the texture of an image \cite{huang2017arbitrary,li2017universal,sheng2018avatar}, the genre or instrument of a music piece \cite{lu2018transferring,lu2018play}, and the sentiment of texts \cite{shen2017style}, with all their content information being preserved. Despite its success, one notable issue in these style transfer methods is that almost all of them operate merely within one single data modality, e.g., from one image to another image, or from one music piece to another. Such a situation is far from the cases that human beings design, create, and interpret an artwork, where good ideas and inspirations usually stem from the interplay among the materials from different data modalities. It is quite natural for human artists to break the restriction of data modality, by projecting their imagination of a music piece into their paintings, or by altering a paragraph of texts in a novel into a scene in a movie. In such processes, one would consider a generation problem mapping from one data modality to another via a latent space, and this latent space properly encodes the styles shared by both sides. This problem is referred to as \emph{cross-modal style transfer} in this paper.

% \begin{figure}[t]
%     \centering
%     \includegraphics[width=0.99\linewidth]{Figure/teaser5.png}
%     \caption{Let music change the visual style of an image. For example, a spectrogram of Claude Debussy's music \emph{Sarabande} in \emph{Pour le piano, L. 95} (1901) transfers Alfred Jacob Miller's painting \emph{The Lake Her Lone Bosom Expands to the Sky} (1850) into an Impressionism-like color scheme through a neural network linking the semantic space shared by music and image.}
%     \label{fig:teaser}
% \end{figure}

Previous investigation of cross-modal learning has been mostly focused on \emph{content generation} rather than style transfer. Most of these studies leverage the techniques of deep transfer learning for various tasks, such as generating images from sound or sounds from images \cite{zhou2018visual,zhao2018sound,wan2019towards}. In comparison to cross-modal content generation, endeavors to cross-modal style transfer are still rarely investigated. However, cross-modal style transfer is often a critical part in the practice of creativity. For example, in design products of virtual reality, animation, and interactive arts, to synergize the styles of visual and music contents is a complicated job, and an automatic process could greatly reduce the efforts. Cross-modal style transfer therefore opens a much broader yet unexplored field for deep generation models. 

% (AAAI, 184 words) 
% In this paper, we for the first time propose a music-to-visual style transfer network. As demonstrated in Figure \ref{figure:teaser}, this problem is considered as an extension of the image style transfer problem, while the style information is provided by music rather than images. We assume that music and image styles can be properly linked by their shared semantic labels; for example, an input image in the left of Figure \ref{figure:GAN} is encoded with a Baroque music piece, and its style becomes similar to Baroque-style paintings in the end, given that the training set contains music and image data both labeled as in the Baroque period. We show that through the proposed \emph{music visualization} network, music can be mapped into image styles in an end-to-end manner. We also show how auxiliary classification and adversarial training with the shared label information improve the training accuracy. Both image and video generation systems are implemented based on our newly proposed dataset, and we discuss the results in two dimensions. One is the compatibility between music and the generated visual contents, and the other is the overall aesthetic experience. 

% (IJCAI, 186 words)
In this paper, we for the first time investigate \emph{music-to-visual style transfer}, a cross-modal style transfer task considered as image style transfer with music as extra condition. This task is conceptually demonstrated in Figure \ref{fig:teaser}, where an Impressionism music piece is encoded to modify the color scheme an American painting in the mid-19th century into an Impressionism-like one. Such utility demonstrates great potential in integrating visual and music contents in animation, virtual reality, and real-world environments such as concerts.

The music-to-visual style transfer network is proposed based on an assumption that music and image styles can be properly linked by their shared semantic labels. With this assumption, three major questions need to be answered: 
1) how to link such semantic labels together,  2) how to evaluate the efficacy of each network component in achieving such aesthetic quality by learning such semantic labels, and 3) how to evaluate the aesthetic quality of the results. We will answer the first question in Section 3 and 4, by introducing the proposed dataset and the music visualization network. Questions 2 and 3 will be discussed in Section 5.

%1) how to link such semantic labels together, and 2) how to evaluate the aesthetic quality of the results, and 3) how to evaluate the efficacy of each network component in achieving such aesthetic quality by learning such semantic labels. We will deal with the first question in Section 3 and 4, by introducing the proposed dataset and the \emph{music visualization} network. The last two questions will be discussed in Section 5.

\section{Related Works}

\subsection{Visual style transfer}

Given a content image and a style image as input, an image style transfer network typically incorporates two tasks: reconstructing the content image or its representation and approximating the statistics of the texture representations (e.g., the Gram matrix) of the style image \cite{gatys2016image}. Notable developments include the Gram matrix of feature maps \cite{gatys2016image}, adaptive instance normalization (AdaIN) \cite{huang2017arbitrary}, whitening and coloring transform (WCT) \cite{li2017universal}, and patch-based methods \cite{sheng2018avatar}. While early developments in style transfer were usually limited by the speed of inference and the classes of output style, recent state-of-the-art style transfer methods have overcome these issues and have shown great potential in online video artistic style transfer \cite{gao2018reconet,chen2017coherent}.

\subsection{Music style transfer}

It is hard to give a holistic definition of music style. Music styles depend on the semantic domain being discussed, such as timbre, performance, or composition styles. Timbre style transfer is usually audio-to-audio style transfer which aims at modifying the timbre such as instrument~\cite{engel2017neural,verma2018neural,lu2018play} or the gender of singers' voice~\cite{kobayashi2014statistical,wu2018singing}. Performance style transfer can be either audio-to-audio or symbolic-to-audio, the latter such as piano performance rendering \cite{hawthorne2018transformer,maezawa2018deep} refers to the tasks of converting deadpan performance data (e.g., MIDI) into expressive performance with a specific interpretation of timing and dynamics. Finally, composition style transfer is usually a symbolic-to-symbolic style transfer problem, which aims at modifying the harmonic, rhythmic or structural attributes of music at the score level, and is applied in music genre transfer \cite{malik2017neural,brunner2018symbolic,lu2018transferring} or blending \cite{kaliakatsos2017conceptual}.  

\subsection{Audio-visual content generation and style transfer}
Deep learning has provided unprecedented flexibility in various cross-modal content generation tasks operating among audio, visual, and textual information. Both audio-to-visual or visual-to-audio content generation has been studied. \cite{zhao2018sound} proposed source separation based on the visual information in music performance. \cite{wan2019towards} proposed an image generation framework taking sound as input. \cite{zhou2018visual} proposed to generate ambient sound or soundscape from a given image. 

Cross-model style transfer is still a rarely investigated topic by now. Recently, \cite{chelaramani2018cross} proposed a text-to-image style transfer framework. Some performance generation works can be regarded as text-to-music style transfer, and one important aim of performance generation is to add expressiveness to a dead-pan performance while preserving its content.
How artists and composers transfer visual contents into musical ideas and vice versa has long been an attractive topic in art studies \cite{berman1999synesthesia,kennedy2007painting}. Continuous efforts for centuries to unravel the relationship between music and visual contents have also engendered new art forms and tools, such as color music \cite{peacock1988instruments}, Lumia arts \cite{collopy2000color} and \emph{music visualization} techniques which are widely seen in a modern multimedia world \cite{mardirossian2007visualizing,bergstrom2007isochords}.

\begin{figure*}[t]
    \centering
    \includegraphics[width=\linewidth]{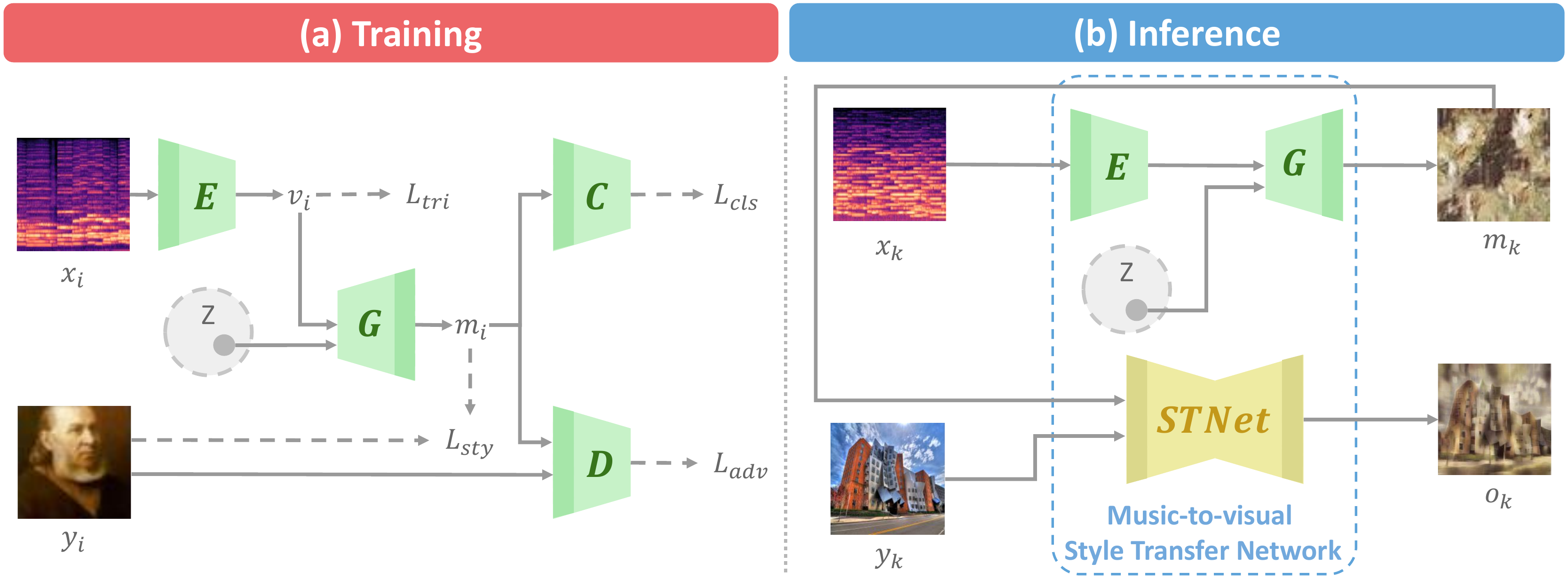}
    \caption{Overview of the music-to-visual style transfer framework. (a) The training scheme of the music visualization net (MVNet), which contains the encoder-generator pair $\{E,G\}$. The classifier $C$ and discriminator $D$ are for regularization and adversarial training. (b) The inference scheme. The trained MVNet can be integrated with an arbitrary style transfer network for image style transfer, based on the visualized music representation outputted by the MVNet.}
    \label{fig:pipeline}
    \vspace{0.2cm}
\end{figure*}

\section{Dataset}

To make a machine learn the relationship between the styles of visual arts and music, we need a dataset containing music and images sharing the same set of semantic labels. How to present such connection between music and images is challenging. The most straightforward way to achieve this goal might be taking the era (i.e., the years the music was composed or the visual artwork was made) as the shared label, as the first attempt to achieve this challenge.\footnote{In this paper, the term \emph{era} may relate to a particular event or movement (e.g., Romanticism), or it may just refer to an arbitrary time interval (e.g., 1700-1710).} Other kinds of labels such as genres, user preference, and emotions are likely to limit the size of the dataset because of the incompatibility of labels in different domains. We consider compiling our dataset from two resources: the WikiArt archive\footnote{\url{https://www.wikiart.org/}} for Western visual arts, and the International Music Score Library Project (IMSLP) online music library\footnote{\url{https://imslp.org/wiki/Main_Page}} for Western classical music, both of which contain era label. For the paintings, we use the \texttt{Wikiart Retriever}\footnote{ \url{https://github.com/lucasdavid/wikiart}} to obtain the images and the corresponding meta-data. As for the music, we use \texttt{Selenium}\footnote{\url{https://pypi.org/project/selenium/}} to retrieve classical music pieces in the IMSLP music library. We choose the data from 1480 to the present and annotate their era labels by decades. For example, the music and paintings in 1700-1710 share one label, and those in 1710-1720 share another label. The proposed dataset, named as the WikiArt-IMSLP dataset hereafter, contains in total 11,127 music pieces and 62,968 paintings divided into 54 classes. To the best of our knowledge, this dataset is the first open-source dataset which pairs the music and visual art together. The dataset will be released after the paper is accepted. 

Note that the labels in the WikiArt-IMSLP dataset are imbalance. %\textcolor{red}{(17.61\% of the images are portrait, 16.70\% landscape, etc.)}%
%\textcolor{red}{(19.96\% of the images are portrait, 17.93\% landscape, etc.)}. 
To facilitate the training process, we choose only portraits (which is the largest category in the WikiArt archive) for training. For music, we choose up to 100 music pieces in each era. As a result, there are 11,078 images and 5,587 music pieces for training. 

\begin{table*}[htbp]
    % \centering
    \caption{The architecture of the networks adopted in this work.}
    \begin{tabular}{ccc}
        %%%%%%%%%%%%%%%%%%%%%%%%%%%%%%%%%%%%%%%%%%%%%%%%%%%%%%%%
        % Generator
        %%%%%%%%%%%%%%%%%%%%%%%%%%%%%%%%%%%%%%%%%%%%%%%%%%%%%%%%
        \vspace{3mm}
        \begin{subtable}{0\textwidth}
            \centering
            {
                \begin{tabular}{@{}cc@{}}
        \toprule
        \midrule
        \multicolumn{2}{c}{$x_i \in{\mathbb{R}^{128 \times 256}} \sim X$} \\
        \midrule
        \begin{tabular}[c]{@{}c@{}} Conv$3\times3$-BN-ReLU \end{tabular}
                     & $128 \times 256 \times 32$ \\
        \midrule
        \begin{tabular}[c]{@{}c@{}} MaxPool-Conv$3\times3$-BN-ReLU \end{tabular}
                     & $ 64 \times 128 \times 64$ \\
        \midrule
        \begin{tabular}[c]{@{}c@{}} MaxPool-Conv$3\times3$-BN-ReLU \end{tabular}
                     & $ 32 \times 64 \times 128$ \\
        \midrule
        \begin{tabular}[c]{@{}c@{}} MaxPool-Conv$3\times3$-BN-ReLU \end{tabular}
                     & $16 \times 32 \times 256$ \\
        \midrule
        \begin{tabular}[c]{@{}c@{}} MaxPool-Conv$3\times3$-BN-ReLU \end{tabular}
                     & $ 8 \times 16 \times 256$ \\
        \midrule
        \begin{tabular}[c]{@{}c@{}} Conv$1\times1$ \end{tabular}
                     & $ 8 \times 16 \times 1$ \\
        \midrule
    \end{tabular}
            } 
        \end{subtable} 
        \hspace{60mm}
        &
        
        \begin{subtable}{0\textwidth}
            \centering
            {
                \begin{tabular}{@{}cc@{}}
        \toprule
        \midrule
        \multicolumn{2}{c}{$\{v_i \oplus z\} \in{\mathbb{R}^{ 8 \times 8 \times 257}}$} \\
        \midrule
        \begin{tabular}[c]{@{}c@{}} DeConv$4\times4$-IN-ReLU \end{tabular}
                     & $16 \times 16 \times 256$ \\
        \midrule
        \begin{tabular}[c]{@{}c@{}} DeConv$4\times4$-IN-ReLU \end{tabular}
                     & $ 32 \times 32 \times 128$ \\
        \midrule
        \begin{tabular}[c]{@{}c@{}} Self attention module \end{tabular}
                     & $32 \times 32 \times 128$ \\
        \midrule
        \begin{tabular}[c]{@{}c@{}} Self attention module \end{tabular}
                     & $32 \times 32 \times 128$ \\
        \midrule
        \begin{tabular}[c]{@{}c@{}} DeConv$4\times4$-Tanh \end{tabular}
                     & $64 \times 64 \times 3$ \\
        \midrule
    \end{tabular}
            } 
        \end{subtable} 
        \hspace{50mm}
        &
        \begin{subtable}{0\textwidth}
            \centering
            {
                \begin{tabular}{@{}cc@{}}
        \toprule
        \midrule
        \multicolumn{2}{c}{$y_i \in{\mathbb{R}^{64 \times 64 \times 3}} \sim Y$} \\
        \midrule
        \begin{tabular}[c]{@{}c@{}} Conv$4\times4$-LeakyReLU \end{tabular}
                     & $32 \times 32 \times 64$ \\
        \midrule
        \begin{tabular}[c]{@{}c@{}} Conv$4\times4$-LeakyReLU \end{tabular}
                     & $16 \times 16 \times 128$ \\
        \midrule
        \begin{tabular}[c]{@{}c@{}} Conv$4\times4$-LeakyReLU \end{tabular}
                     & $8 \times 8 \times 256$ \\
        \midrule
        \begin{tabular}[c]{@{}c@{}} Conv$4\times4$-LeakyReLU \end{tabular}
                     & $4 \times 4 \times 256$ \\
        \midrule
    \end{tabular}
            } 
        \end{subtable} 
        \hspace{50mm}
        \\
        \vspace{3mm} (a) Encoder  & (b) Generator \hspace{0mm} & (c) Discriminator\\
    \end{tabular}
    \label{table::arch}
\end{table*}

\section{Music-to-Visual Style Transfer}

We solve the music-to-visual style transfer problem with two steps, namely music visualization and style transfer.
Figure~\ref{fig:pipeline} illustrates the pipeline of the proposed system. The system contains two major networks, which are referred to as the \emph{music visualization net} (MVNet) and the \emph{style transfer net} (STNet) in this paper. The MVNet is a regularized encoder-decoder network; its input is an audio data representation, and its output is an image which resembles the style of that image paired with the audio. This image will be referred to as the \emph{style image} hereafter. The style image generated by the MVNet and the target image (i.e. content image) are then fed into the STNet. The output of the STNet is a modified image which resembles the style of the style image.

In what follows, we consider the training data with $N$ music-image pairs $\{x_i, y_i\}^N_{i=1}$, where the $x$ being the 2-D mel-spectrogram of music signals and $y$ being the corresponding image, such that each $x_i$ and $y_i$ were created in the same era. For simplicity, the dimension of all the images is adjusted to $64 \times 64$. The music signals are clips with the length of 8.91 seconds, segmented at the first one-third of each music piece in the dataset. The sampling rate of the music signals is 22.05 kHz. Hamming window with the size of 1024 and hop size of 256 are used for computing spectrogram. The size of the mel-filterbank is 128. The mel-spectrogram is divided into three parts, each of which with 2.97 seconds length. The three parts are then assigned to the three input channels, resulting in the dimension of $128 \times 256 \times 3$.
% \footnote{Our pilot study showed that a single-channel feature is insufficient in representing music information possibly because it is too short. Considering a longer piece of music could solve this issue, but in this case, the input dimension is too large and the model becomes unstable to converge. One compromise is to use a long music segment but split it into three channels. In this case, the role of the convolutional filters in the network would differ from the usual ones in processing the RGB channels in images, as now the mel-spectrograms in the three channels are not necessarily synchronized.} 
The mel-spectrogram is obtained with the \texttt{librosa} library.\footnote{\url{https://librosa.github.io/librosa/}}

The main reason that we divide a mel-spctrogram into three channels for training rather than using merely one channel is described as follows. Our pilot study showed that a single-channel feature is insufficient in representing music information possibly because it is too short. Considering a longer piece of music could solve this issue, but in this case, the input dimension is too large and the model becomes unstable to converge. One compromise is to use a long music segment but split it into three channels. In this case, the role of the convolutional filters in the network would differ from the usual ones in processing the RGB channels in images, as now the mel-spectrograms in the three channels are not necessarily synchronized.

\subsection{Training the Music Visualization Net (MVNet)}

The left part of Fig. \ref{fig:pipeline} demonstrates the MVNet in the training stage, which contains an encoder $E$, a generator $G$, a discriminator $D$, and an auxiliary classifier $C$. The encoder $E$ first encodes $x_i$ into a latent vector $v_i:=E(x_i)$. A triplet loss term $L_{tri}$ \cite{schroff2015facenet} regularizes the behaviors of $v_i$ such that any two $v_i$s encoded from the music pieces in the same era are as similar to each other as possible, while any two $v_i$s from different era are as far to each other as possible. That means, for every input triplet $(x^a_i, x^p_i, x^n_i)$ where $x^p_i$ and $x^a_i$ are in the same era, $x^n_i$ and $x^a_i$ are in different eras, and $(v^a_i, v^p_i, v^n_i)=(E(x^a_i), E(x^p_i), E(x^n_i))$, the triplet loss is represented as
\begin{equation}
    L_{tri} = \sum^N_{i=1} \left[\| v^a_i - v^p_i \|^2_2 - \| v^a_i - v^n_i  \|^2_2 + \alpha \right]_+
\end{equation}
where $\alpha$ is the margin which is set to $1.0$ in our experiment, and $[\cdot]_+:=\max(0,\cdot)$. In the training process, we select $x^p_i$ and $x^n_i$ by randomly sampling images from training set.

The latent vector $v_i$ is then concatenated with a random vector $z\sim\mathcal{N}(0, I)$ and fed into the generator $G$. The output $m_i$ is represented $m_i=G(v_i \oplus z)$. To make $m_i$ resemble the style of $y_i$, we impose a style loss term $L_{sty}$ on the generator network. The style loss is defined as the $l_1$ distance in Gram matrices between two images, and we follow the method in \cite{gatys2016image} to compute the style loss based on the feature maps of a pre-trained VGG net \cite{simonyan2014vgg}:  
\begin{equation}
L_{sty} = \sum^N_{i=1}\sum^S_{s=1} \lVert Gram(VGG_s(m_i)) - Gram(VGG_s(y_i)) \rVert_1
\end{equation}
where $S$ is the number of layers, $Gram (\cdot)$ is the Gram matrix operation \cite{gatys2016image}, and $VGG_s(\cdot)$ represents the $s$th-layered feature map of the pre-trained VGG net.

We also introduce an adversarial loss $L_{adv}$ to enhance the training process. Following \cite{lu2018play}, we employ RaGAN~\cite{jolicoeur2018relativistic} as our adversarial training mechanism. Let $\bar{D}$ represent the discriminator $D$ without the last sigmoid layer; that means, $D(a, b):=\text{sigmoid}(\bar{D}(a) - \bar{D}(b))$. Let $P_{real}$ and $P_{gen}$ be the distributions of the image and the music representation, respectively. The adversarial loss is represented as 
\begin{equation}
    \begin{split}
        L_{adv} = &-\mathop{\mathbb{E}_{y_i\sim P_{real}}}[\log(\mathop{\mathbb{E}_{m_i\sim P_{gen}}}[D(y_i, m_i)])] \\
        &- \mathop{\mathbb{E}_{m_i\sim P_{gen}}}[\log(1-\mathop{\mathbb{E}_{y_i\sim P_{real}}}[D(m_i, y_i)])]
    \end{split}
    \label{eq::adv}
\end{equation}

Finally, to better utilize the era labels in the training data, we further introduce a classifier $C$ for era classification. By using this classifier we expect that the likelihood function of the music representation $P(c|m_i)$ should approximate the likelihood function of the training image $P(c|y_i)$, where $c$ represents the era labels. We therefore consider the era classification loss $L_{cls}$ for $y_i$ and $m_i$:
\begin{equation}\label{eq::cls_C}
    L_{cls}^C = \sum^N_{i=1} \log P(c | y_i) 
\end{equation}
\begin{equation}\label{eq::cls_G}
    L_{cls}^G = \sum^N_{i=1} \log P(c | m_i)
\end{equation}
where the superscripts $C$ and $G$ indicate the sub-network being updated when that loss term is used during training. More specifically, when training the auxiliary classifier $C$, we adopt $L_{cls}^C$ in Equation~\ref{eq::cls_C} to fit the training image $y_i$ their labels; when training $G$, we adopt $L_{cls}^G$ in Equation~\ref{eq::cls_G} to fit the generated music representation $m_i$ to the label of their corresponding music piece $x_i$. In this way, the generator is regularized by this loss term since $x_i$ and $y_i$ are paired with the same set of era labels.

\begin{figure*}
    \centering
    \includegraphics[width=\linewidth]{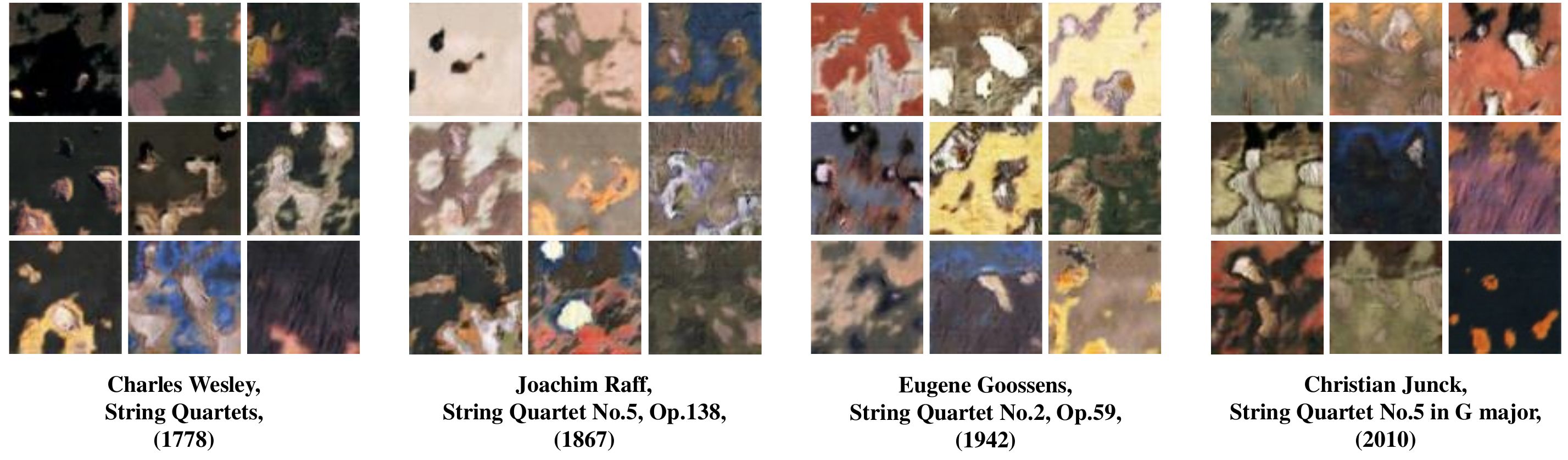}
    \caption{Random music representation samples generated from four string quartets composed in different eras. The composers' name, the title of music, and the year of composition are listed below the generated samples.}
    \label{fig:random_sample}
\end{figure*}

In summary, the total loss function $L$ for training the network $\{E, G, D\}$ is represent as
\begin{equation}
	L = L_{adv} + \alpha L_{tri} + \beta L_{sty} + \gamma L_{cls}^G\,,
	\label{eq::total_loss}
\end{equation}
and the era classifier $C$ is trained solely with $L_{cls}^C$. In our experiment, we set $\alpha=0.1$, $\beta=10.0$ and $\gamma=0.1$. In the following, we refer to the trained encoder-decoder pair $\{E, G\}$ as the MVNet. 

Note that the training pair $\{x_i, y_i\}$ is not uniquely defined. One $x_i$ can be paired with multiple $y_i$ having the same era label to $x_i$. Therefore, to include more possible $\{x_i, y_i\}$ pairs, these pairs are randomly shuffled for every training epoch.

Table~\ref{table::arch} shows the components of the encoder, generator and discriminator. Each row of the tables lists the operations employed in each layer (left, different operators are separated with hyphen), and the size of output feature (right). %, and the number following the convolution is the kernel size. 
The kernel size for each max pooling operation is two. The structure of the encoder is similar to the encoder part of a traditional U-Net \cite{ronneberger2015u}. The architecture of generator generally follows the self-attention GAN (SAGAN)~\cite{zhang2018self}, where we employ the self-attention module after the stacked deconvolution layers. For each deconvolution operator in the generator, we adopt spectral normalization to guarantee the training stability. It should be noted that the music latent vector $v_i$ is resized to $8 \times 8 \times 1$ with bi-linear interpolation, in order to guarantee the consistency among feature dimensions, since the size of the random vector $z$ is $8 \times 8 \times 256$. As a result, the overall size of the feature $\{v_i\oplus z\}$ is $ 8 \times 8 \times 257$. At last, to enforce better modeling of high-frequency structure, we follow the idea of PatchGAN~\cite{zhu2017unpaired}, and have the discriminator output a feature map with size of $4 \times 4 \times 256$.

\subsection{Inference}

The right part of Figure ~\ref{fig:pipeline} demonstrates the inference procedure. After finishing training, the trained MVNet $\{E,G\}$ is employed to generate images (i.e. music representation $m_i$) from music, and the images are expected to convey visual style information of the paintings coming from the era of that music piece.
 
This style image and the content image are fed into the STNet, which can be an arbitrary image style transfer network, such as those being reviewed in the previous sections. More specifically, for a given content image $y_k$ and an arbitrary music segment $x_k$, the stylized image $o_k$ can be generated as:
\begin{equation}
    o_k = STNet(y_k, G(E(x_k)))\,.
\end{equation}

A detailed comparison of various STNets can be found in the next section.

\begin{figure*}
    \centering
    \includegraphics[width=\linewidth]{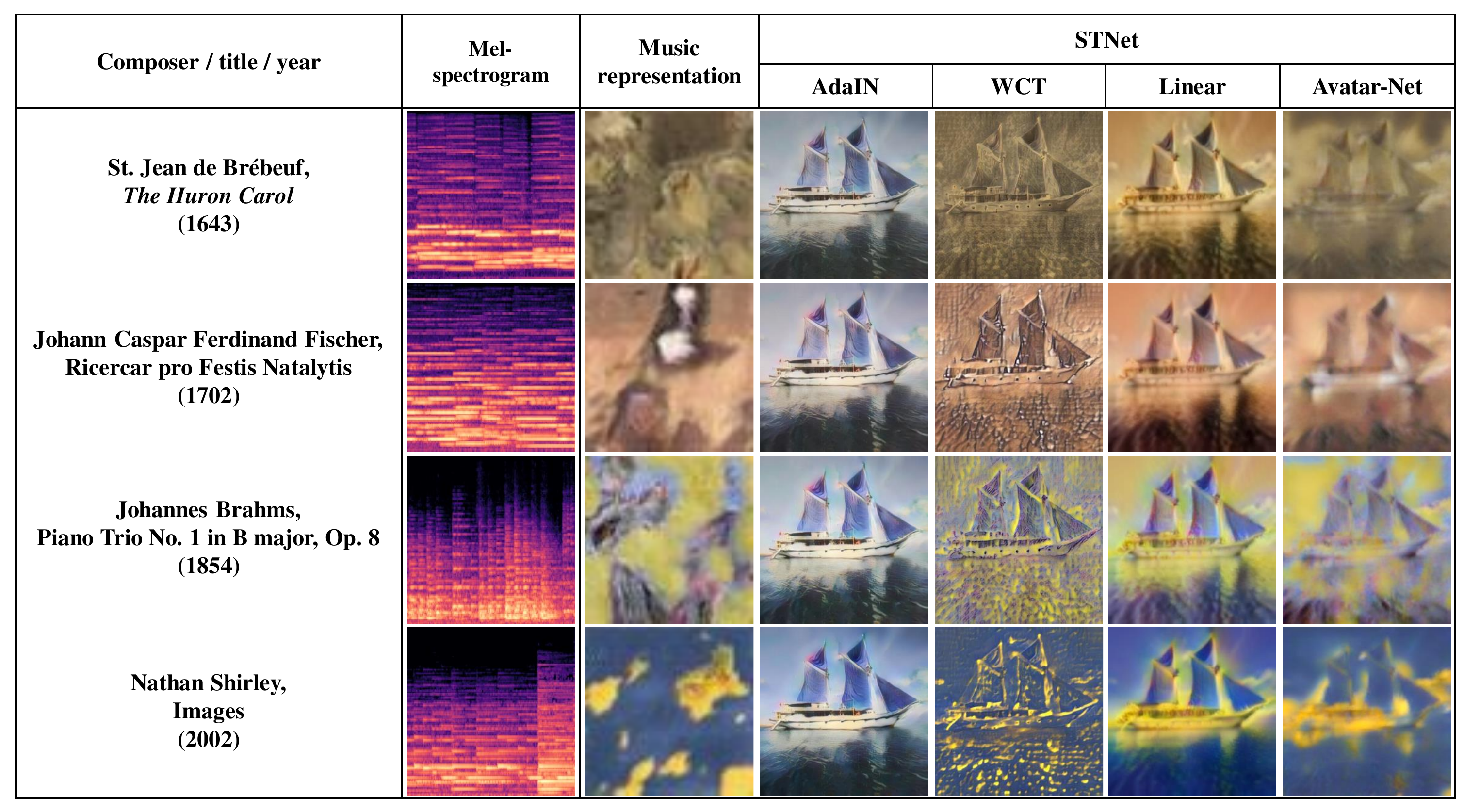}
    \caption{Comparison of music-to-image style transfer results over four music pieces from different eras and four STNets. The original content image can be seen in the left of Figure \ref{fig:teaser}. The first column shows the name of the composer, the title of music, and the year of composition. The mel-spectrograms of the music are also illustrated for reference. More results with different content images are provided in supplementary material. }
    \label{fig:style_transfer_demo}
    \vspace{-2mm}
\end{figure*}

\begin{figure}[t]
    \centering
    \includegraphics[width=\linewidth]{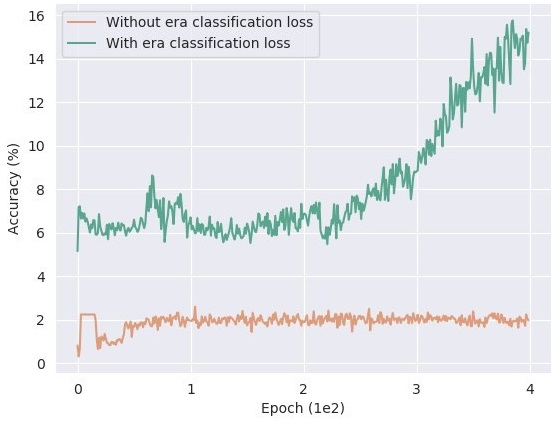}
    \caption{The accuracy of era classification for every epoch.  Green line: with classification loss. Orange line: without era classification loss.}
    \label{fig:acc_curve}
    \vspace{-2mm}
\end{figure}

\section{Experiment and Result}

The model is trained using two GTX-1080 Ti GPUs and 2 TB SSD. Training the MVNet takes around 48 hours to acheive convergence. The system is implemented with Python3.6 and PyTorch deep learning framework. For network optimization, we use the Adam optimizer ($\beta_1 = 0.9$, $\beta_2 = 0.999$), a fixed learning rate $0.0001$, and a batch size $16$ to train the network. Weight decay is set to $0.0001$ to avoid over-fitting. Before the painting images are fed into the network, we resize the image as $96 \times 96$, and randomly crop it with a $64 \times 64$ patch.

Experimental results are demonstrated as follows: First, generated samples of the music representation are illustrated. Second, the performance of the system is assessed through the accuracy of era classification, Third, transferred image and audio samples conditioned on music in different eras and processed with different STNets are illustrated and discussed. Finally, we conduct a systematic user study to evaluate the aesthetic quality of the music-to-visual style transfer system. The supplementary materials, demo images and videos, data, and codes are available at the project webpage: https://sunnerli.github.io/Cross-you-in-style/

\subsection{Illustration of music representations}

To see how the music representations look like, in Figure~\ref{fig:random_sample} we illustrate four sets of music representations, which are generated from four string quartets composed in the years of 1778, 1867, 1942, and 2010, respectively. First, we investigate whether the MVNet can generate diverse outputs by random sampling over $z\sim\mathcal{N}(0, I)$. To verify this, nine samples are selected for each set. The illustrated samples indicate that the MVNet does generate diverse outputs, as none of them look the same as others.

Second, by fixing the inputs to be string quartets, Figure \ref{fig:random_sample} allows us to compare how the \emph{styles} in both music and paintings affect each other through neural network mapping. The compositional styles of the four music pieces are quite different: Wesley is in the Classical period; Raff's work is mixed with the Romantic penchant where transposition and chromatic scales are more actively used; and Goosens and Junck are more or less influenced by post-tonal music and electronic music. These difference can be observed from their mel-spectrograms. The styles of paintings in the four periods are also different; for example, before the mid-19th century, low color saturation and unified color tone are preferred, while after the mid-19th century, colorful elements (e.g., orange, pink, etc.) with high saturation and complementary colors (e.g., orange and blue, yellow and purple) are characteristic.

We observe that what Figure \ref{fig:random_sample} shows is consistent with the aforementioned historical statements on music and arts. The images of the first set (Wesley's string quartet, 1778) are more similar to each other than the other three sets in terms of tone and saturation. The use of complementary colors within one image is also rarely seen in the first set but commonly seen in the other three. This demonstrates that the network does capture the music and image styles and map them to each other properly.
% More examples can be found in Figure \ref{fig:style_transfer_demo} and in the supplementary material\footnote{The link of supplementary material: \url{https://bit.ly/2uijusm}}.
More examples of the music representation can be found in Figure \ref{fig:style_transfer_demo} and the project webpage.\footnote{The link of project page: \url{https://sunnerli.github.io/Cross-you-in-style/}}

% Fig 6
\begin{figure*}
\centering

\begin{tabular}[t]{c@{\hspace{0.01\linewidth}}c@{\hspace{0.01\linewidth}}c@{\hspace{0.01\linewidth}}c@{\hspace{0.01\linewidth}}c@{\hspace{0.01\linewidth}}c@{\hspace{0.01\linewidth}}c@{\hspace{0.01\linewidth}}c}
{Music Title} & {Content image} & {Result (linear)} & {Content image} & {Result (linear)} & {Content image} & {Result (linear)} \\

{}&
\multirow{5}{*}{\includegraphics[width = .14\linewidth]{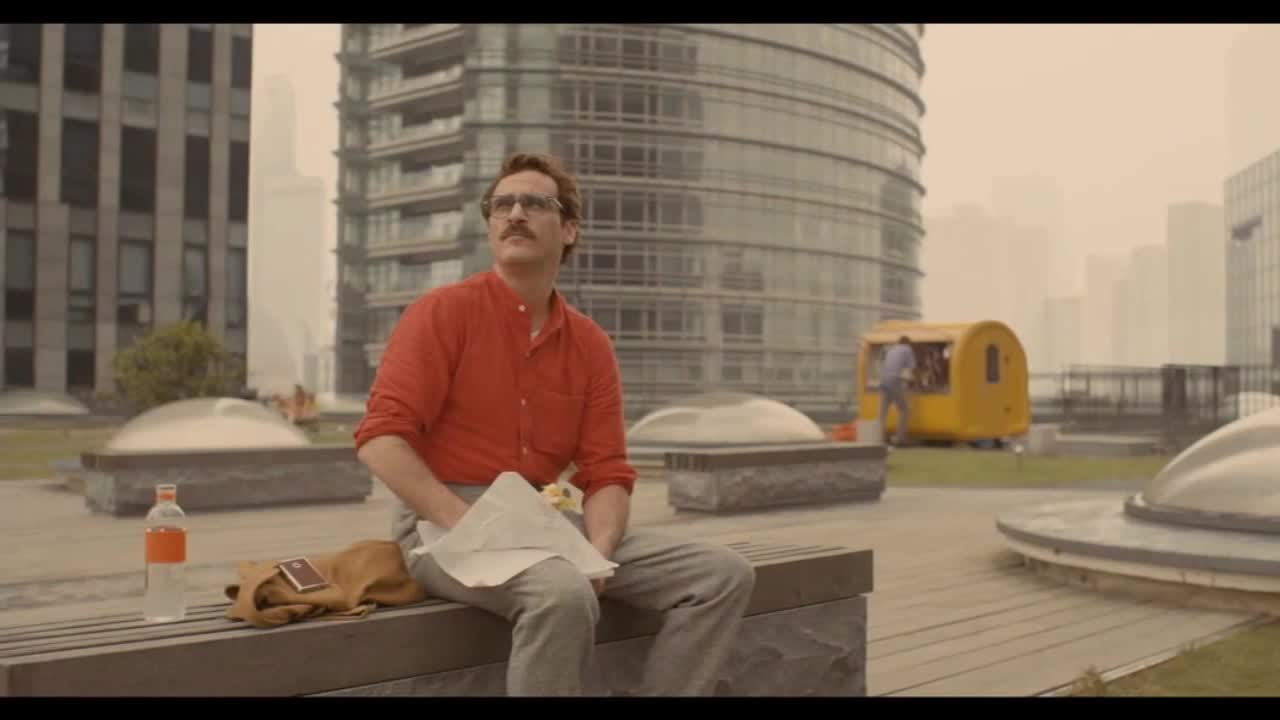}} &
\multirow{5}{*}{\includegraphics[width = .14\linewidth]{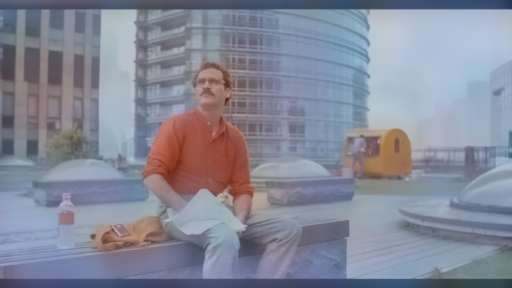}} &
\multirow{5}{*}{\includegraphics[width = .14\linewidth]{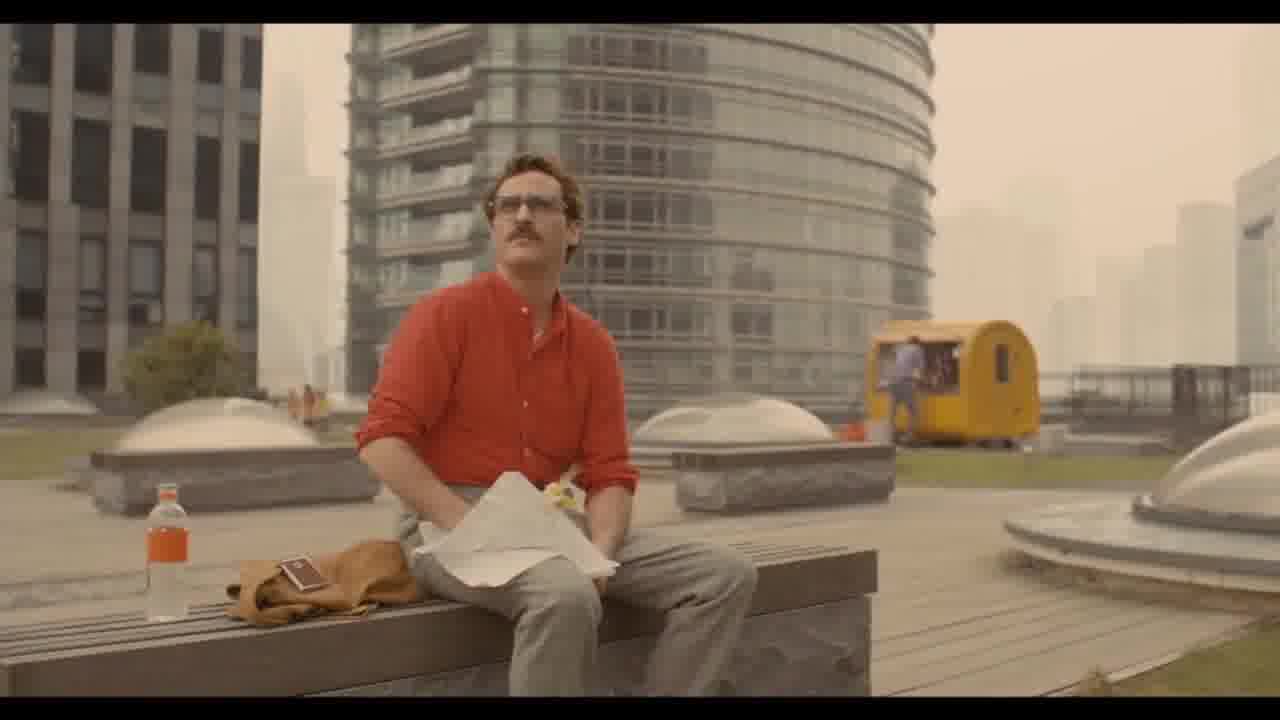}} &
\multirow{5}{*}{\includegraphics[width = .14\linewidth]{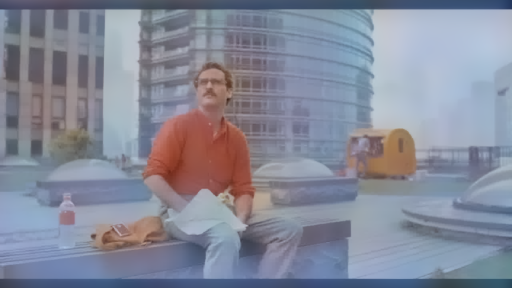}} &
\multirow{5}{*}{\includegraphics[width = .14\linewidth]{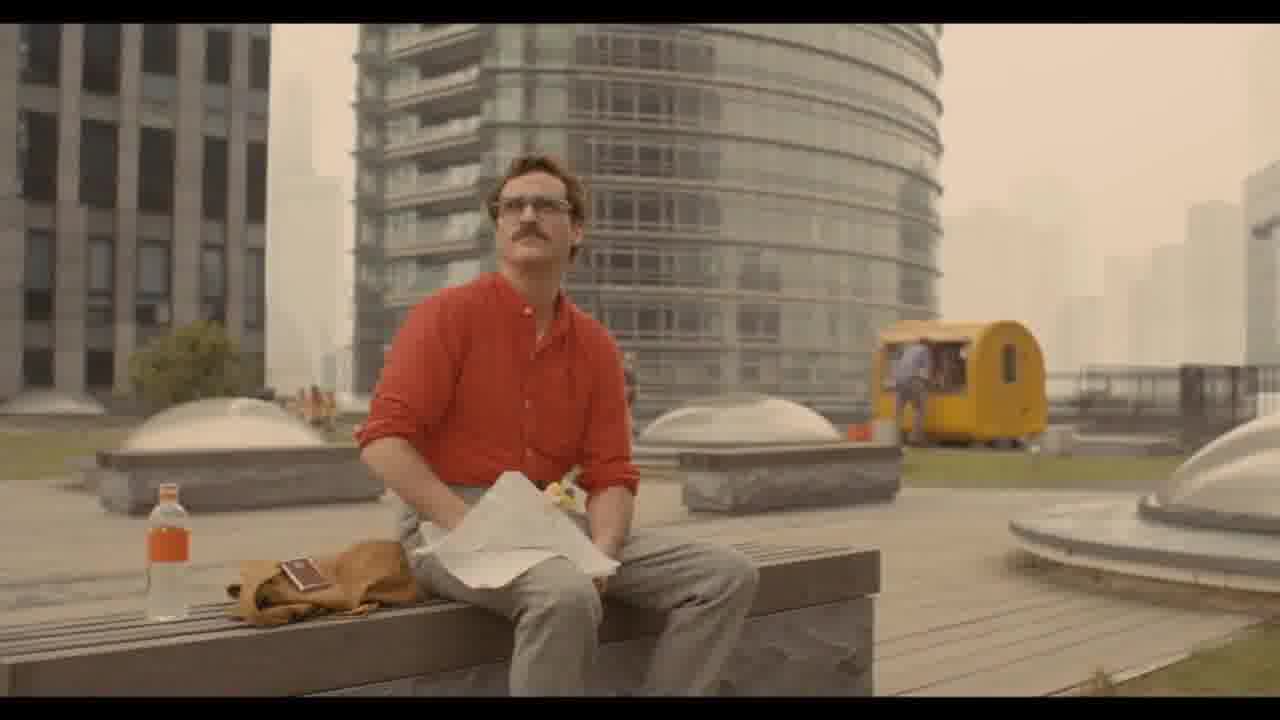}} &
\multirow{5}{*}{\includegraphics[width = .14\linewidth]{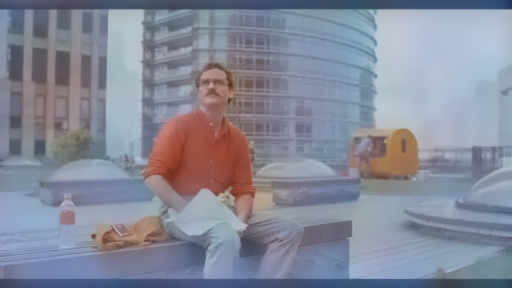}} & \\
\footnotesize Arcade Fire, &&\\
\footnotesize \textit{Photograph},&&\\
\footnotesize (2013)&&\\
% &&\\

{}&
\multirow{5}{*}{\includegraphics[width = .14\linewidth]{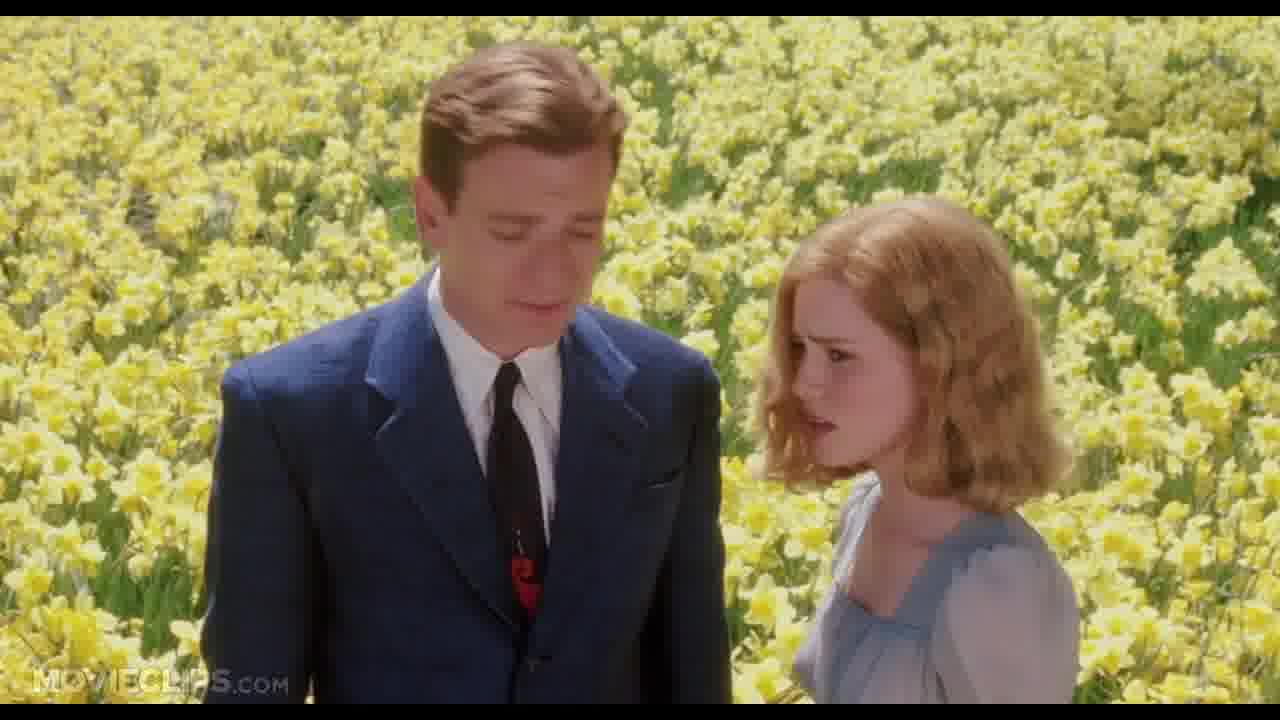}} &
\multirow{5}{*}{\includegraphics[width = .14\linewidth]{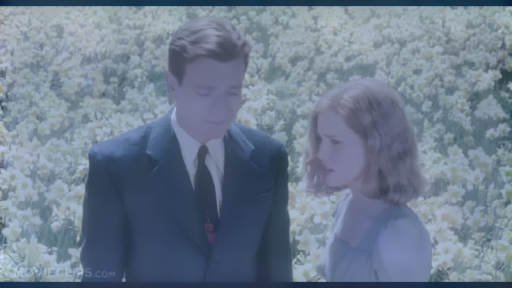}} &
\multirow{5}{*}{\includegraphics[width = .14\linewidth]{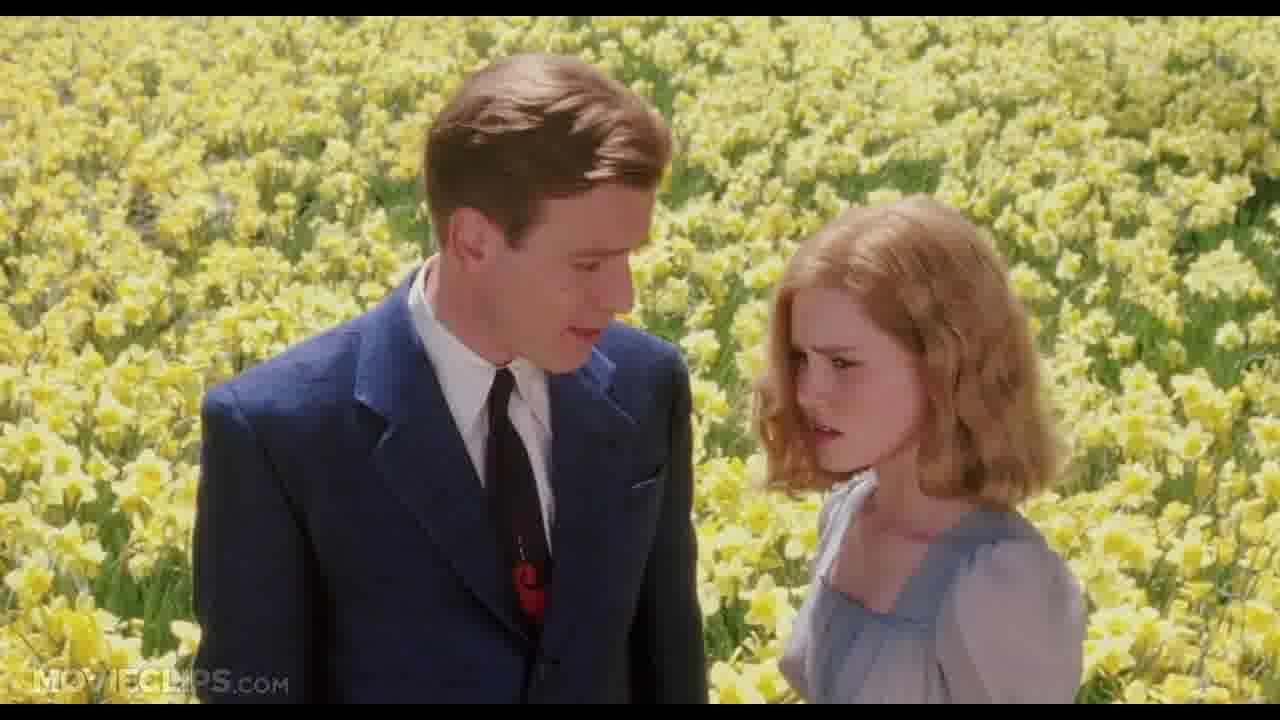}} &
\multirow{5}{*}{\includegraphics[width = .14\linewidth]{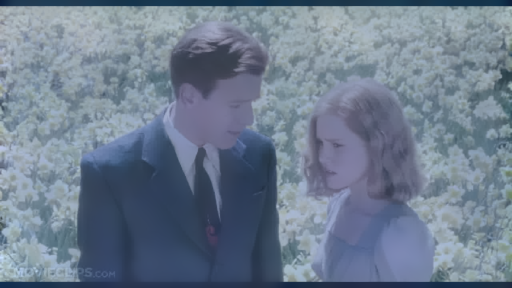}} &
\multirow{5}{*}{\includegraphics[width = .14\linewidth]{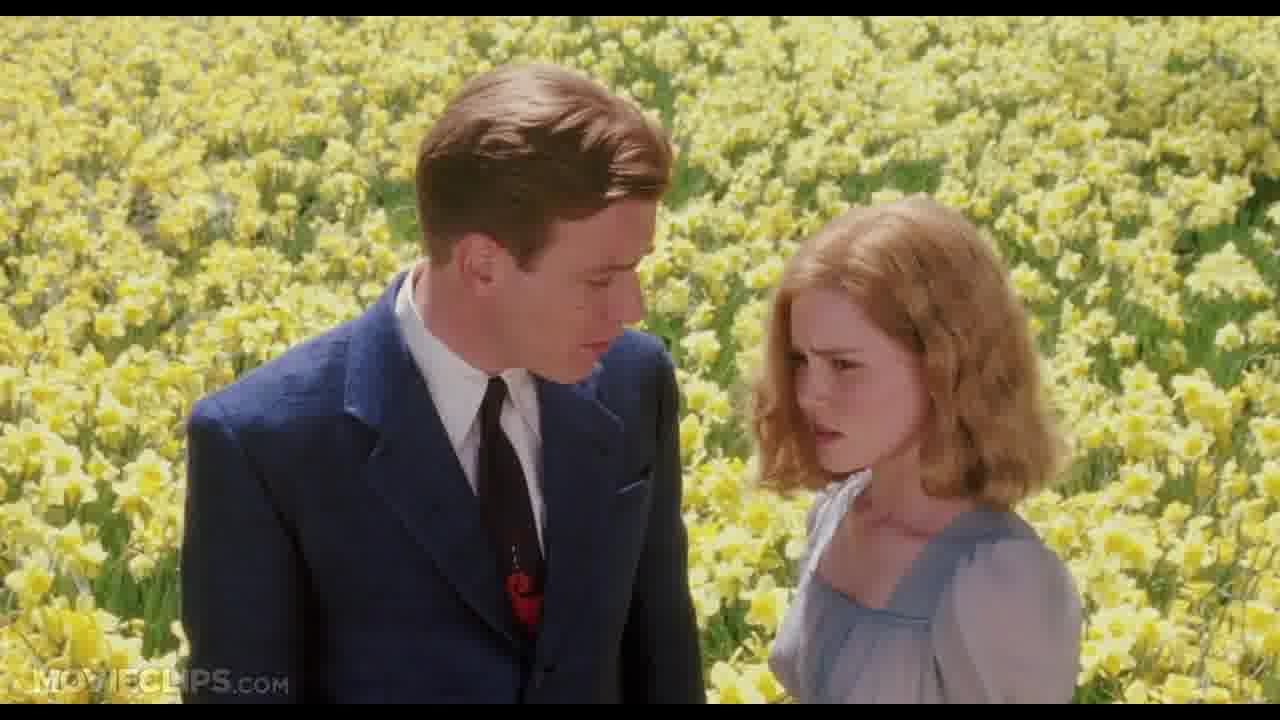}} &
\multirow{5}{*}{\includegraphics[width = .14\linewidth]{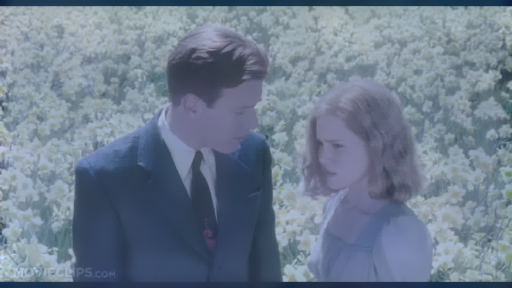}} & \\
\footnotesize Danny Elfman, &&\\
\footnotesize \textit{Sandra's Theme},&&\\
\footnotesize (2003)&&\\
% &&\\

{}&
\multirow{5}{*}{\includegraphics[width = .14\linewidth]{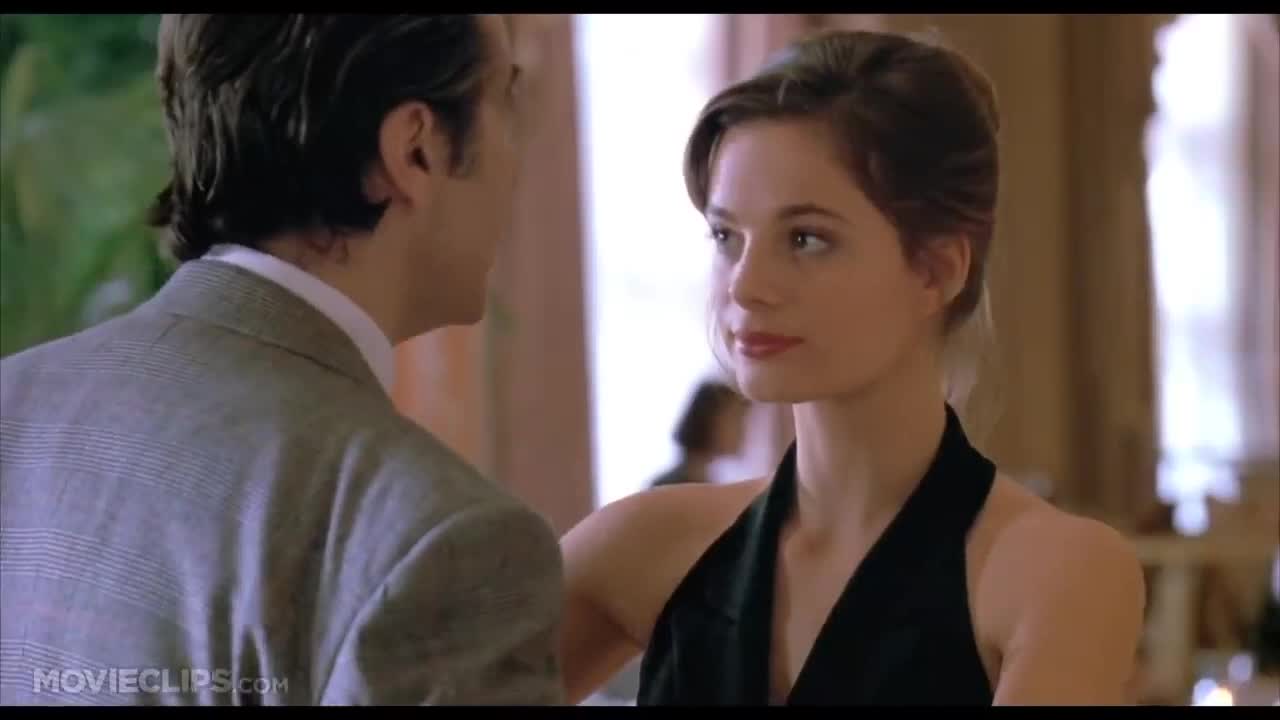}} &
\multirow{5}{*}{\includegraphics[width = .14\linewidth]{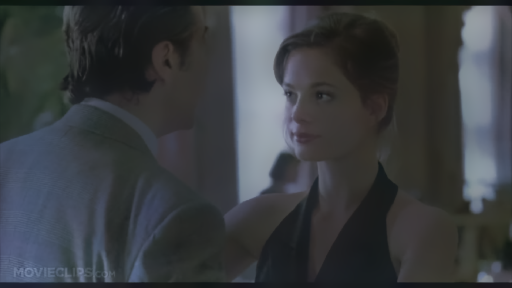}} &
\multirow{5}{*}{\includegraphics[width = .14\linewidth]{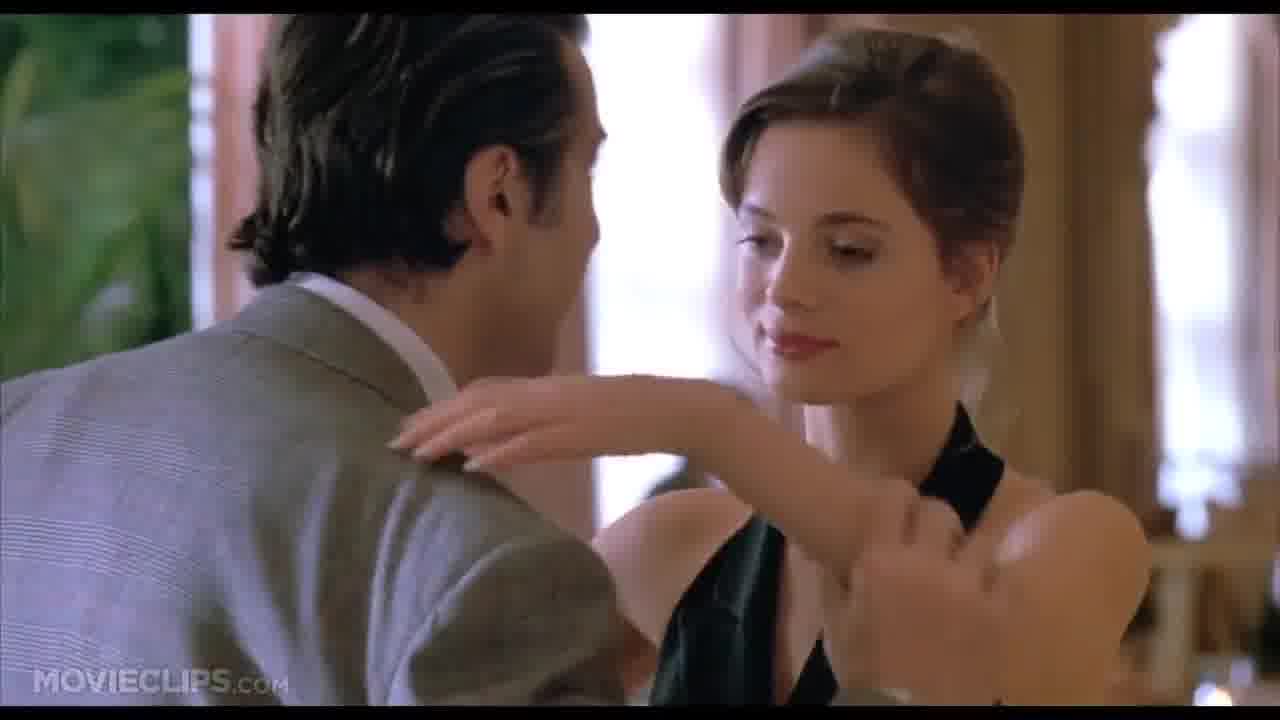}} &
\multirow{5}{*}{\includegraphics[width = .14\linewidth]{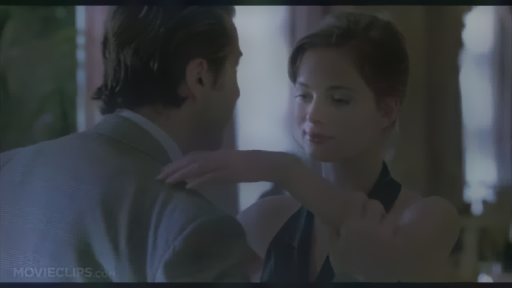}} &
\multirow{5}{*}{\includegraphics[width = .14\linewidth]{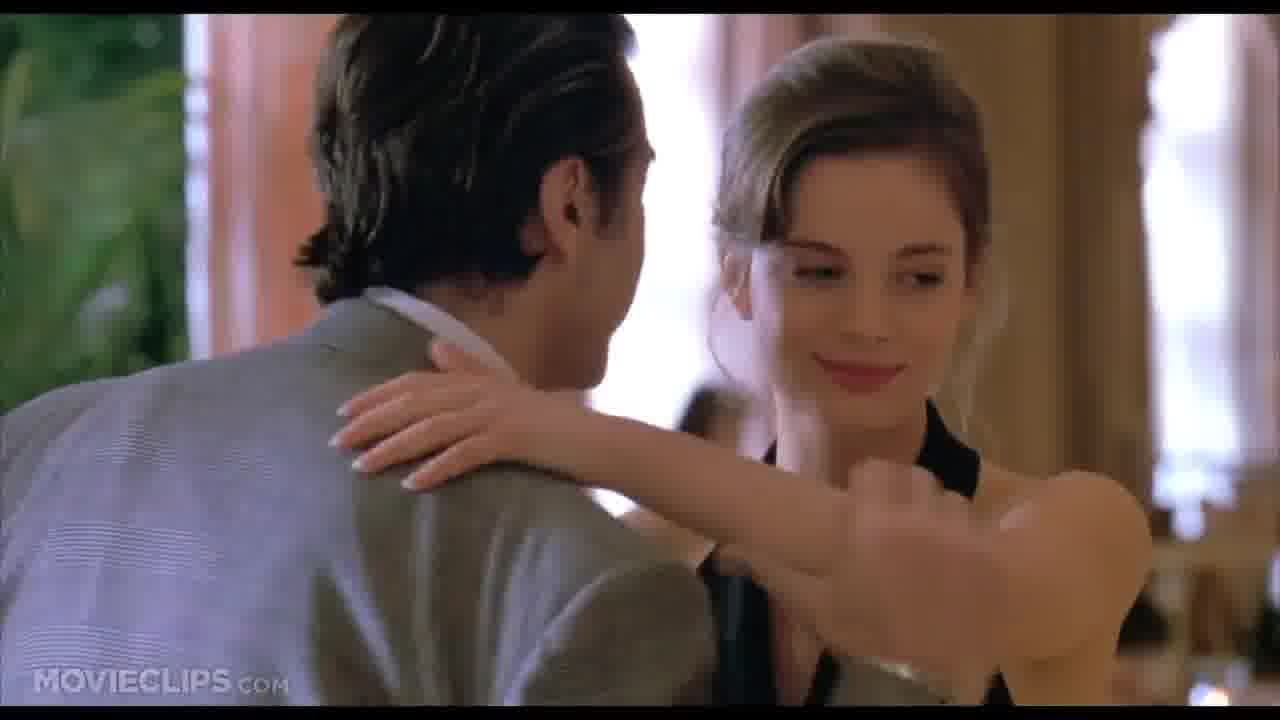}} &
\multirow{5}{*}{\includegraphics[width = .14\linewidth]{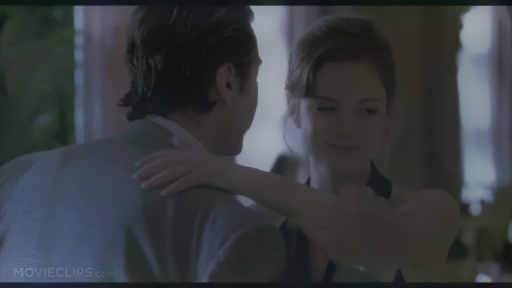}} & \\
\footnotesize Carlos Gardel, &&\\
\footnotesize \textit{Por Una Cabeze},&&\\
\footnotesize (1935)&&\\
% &&\\

{}&
\multirow{5}{*}{\includegraphics[width = .14\linewidth]{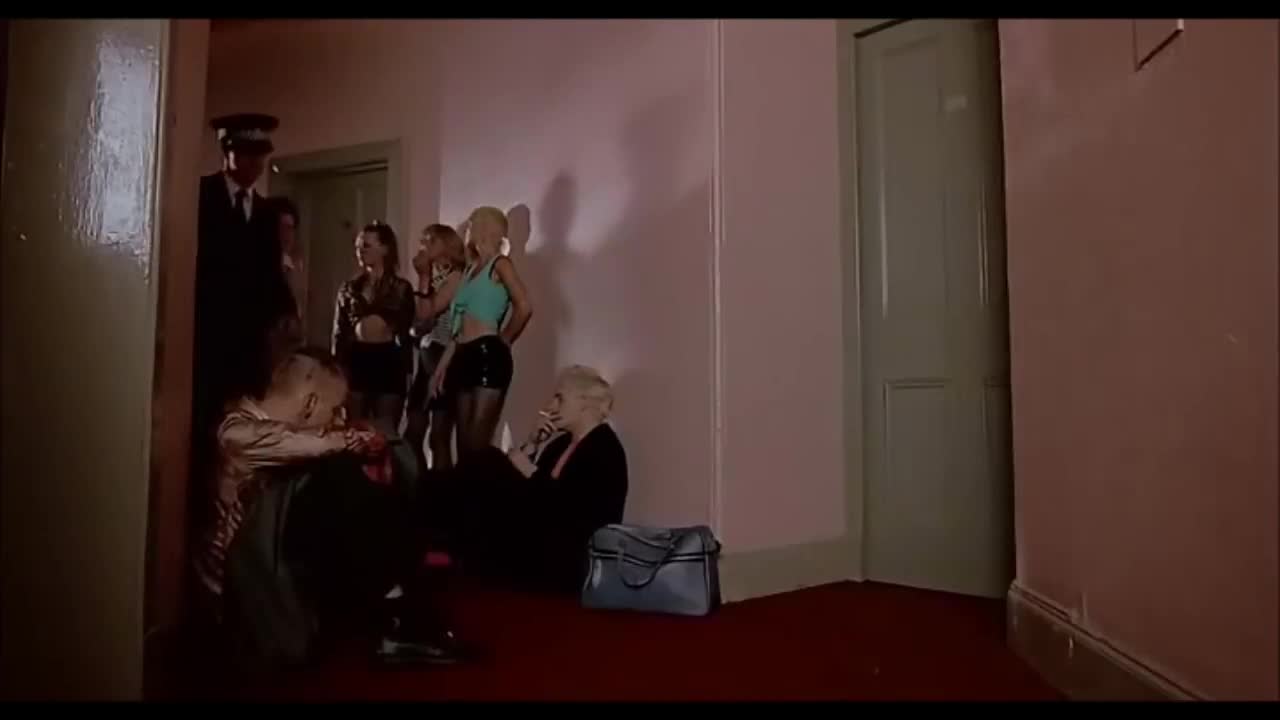}} &
\multirow{5}{*}{\includegraphics[width = .14\linewidth]{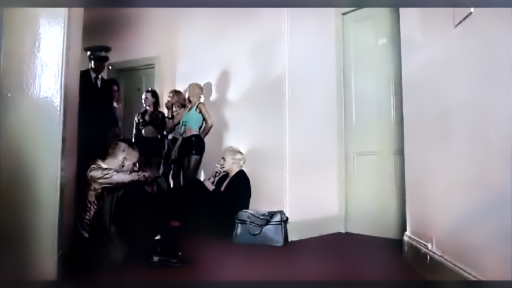}} &
\multirow{5}{*}{\includegraphics[width = .14\linewidth]{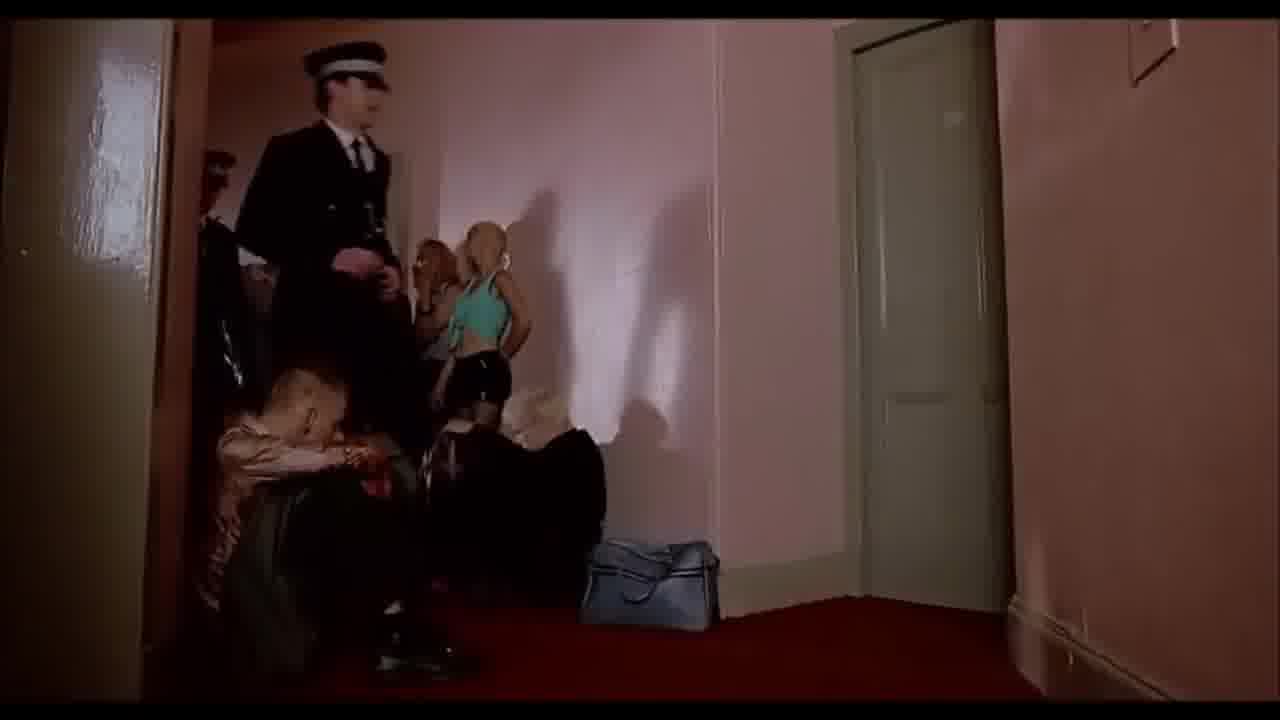}} &
\multirow{5}{*}{\includegraphics[width = .14\linewidth]{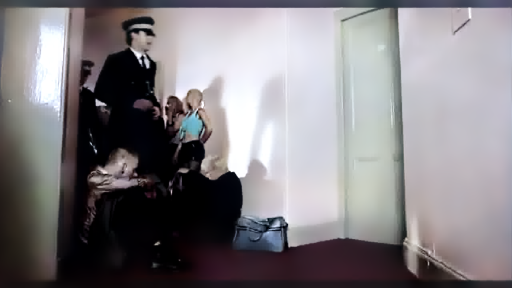}} &
\multirow{5}{*}{\includegraphics[width = .14\linewidth]{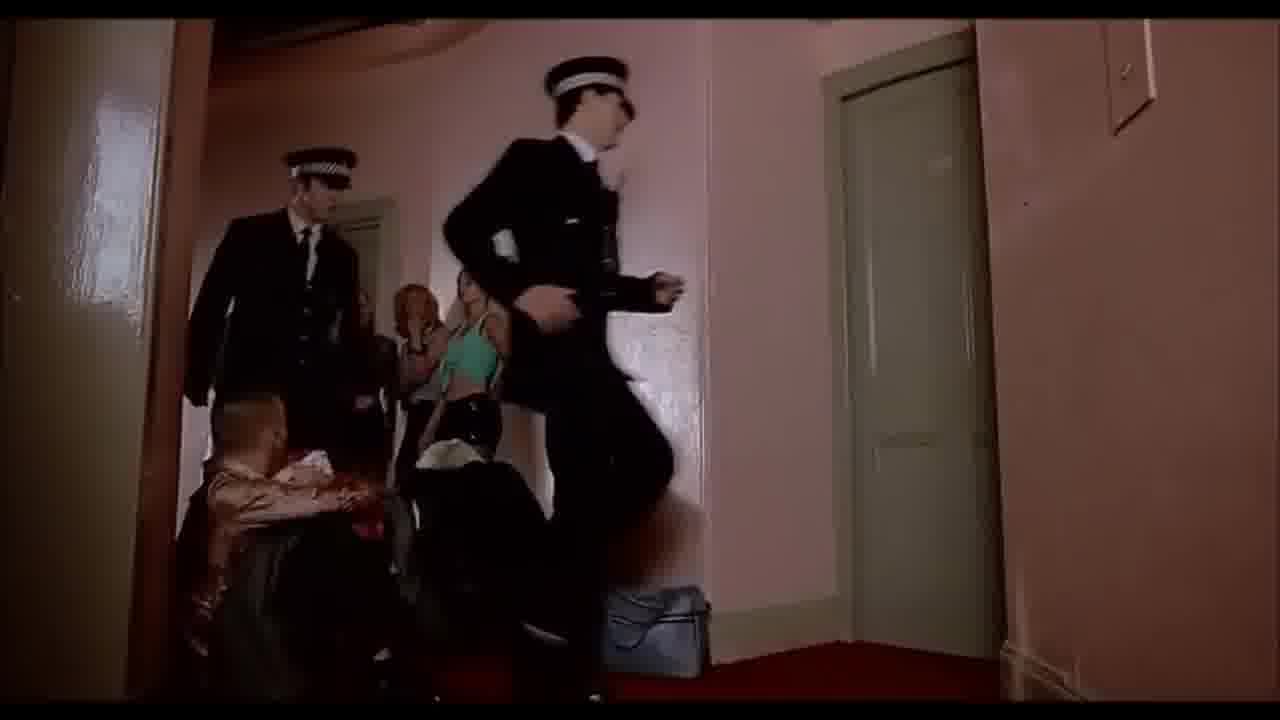}} &
\multirow{5}{*}{\includegraphics[width = .14\linewidth]{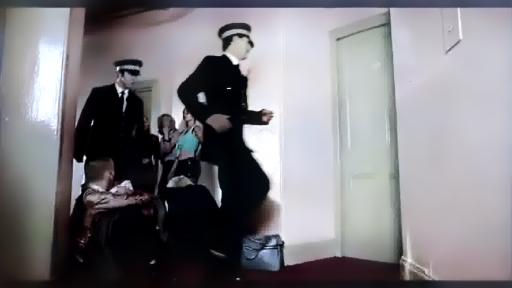}} & \\
\footnotesize Underworld, &&\\
\footnotesize \textit{Born Slippy},&&\\
\footnotesize (1995)&&\\
&&\\

\end{tabular}

\caption{Results of music-to-video transfer. 
% single frame.
The original content images and the transferred results of three selected frames for each video are shown. Music genres from top to down: piano solo, symphony, chamber music, and progressive house. 
% From left to right: music name, content image, and transferred result (one selected frame) using the linear transformation method.
From left to right: composer/ music title/ years, and three pairs of original content image with transferred result which using the linear transformation method.
}
\label{fig:Music2Video_2C}
\end{figure*}
% Fig6

\subsection{Era classification}

Figure~\ref{fig:acc_curve} shows the accuracy on era classification computed over the generated music representations for every epoch. Since there are 54 era classes, the accuracy of a random guess is around 2\%. Two settings are compared in this experiment: the first setting imposes the classifier loss ($\mathcal{L}^G_{cls}$) when training $G$, and the second setting does not include this loss term. The result shows that if the classification loss is not imposed, the accuracy remains at the random guessing level over the epochs. When the classification loss function is imposed, the accuracy increases over the epochs. An increased accuracy implies a higher probability to generate images that can be classified to the correct era through this classifier.

\subsection{Comparison of STNets}\label{discussion::style}

To demonstrate the compatibility of the proposed framework with various style transfer methods, we compare the style transfer outputs generated by four different STNets. The STNets include AdaIN \cite{huang2017arbitrary}, WCT~\cite{li2017universal}, linear transformation~\cite{li2018learning} and Avatar-Net~\cite{sheng2018avatar}. Figure~\ref{fig:style_transfer_demo} shows the input mel-spectrograms, the music representations, and the generated results of the four STNets conditioned on four music pieces in different eras. Results show that, again, the generated music representations do correspond with the painting styles in that era: low saturation and unified hue before the Classic period, while high saturation and complementary colors after Romanticism.

For the transferred results, we found that AdaIN merely captures the colors in the music representations, though its processing speed is the fastest among all. The other three STNets can better capture the color scheme of the music representations. The linear transformation can even capture the brushstroke-like texture. WCT tends to emphasize the boundary on a small scale, and Avatar-Net tends to blur the content image. In summary, we indicate that the linear transformation is a compromise between speed and quality. As a result, we will use the linear transformation method in the remaining experiments.

\subsection{Music-to-video style transfer}

In the above discussion, a music segment is assumed to be a static object. We then consider a more realistic case that how music, as an art of time, modifies the visual styles of video in a dynamical manner. As a proof-of-concept study, we consider a preliminary scenario: we select movie clips with background music arrangement, resample the clips to 20 fps, and then for every video frame we take the background music segment around the video frame as the music input to transfer the visual style of that video frame. The length of the background music segment is 8.91 secs and the middle of this segment is at the time of the video frame. That means, style transfer is processed frame by frame, and each frame takes different but overlapped music segments to generate the music representations. For simplicity, the latent vector $z$ is kept the same over time. 
We selected four movie clips from \emph{Her} (2013), \emph{Big Fish} (2003), \emph{Scent of a Woman} (1992), and \emph{Trainspotting} (1996) whose background music are piano solo, symphony, chamber music, and progressive house, respectively. The last one does not belong to the classic repertoire. The titles of the music and the composers' name are listed in Figure \ref{fig:Music2Video_2C}. 
The movie clips were retrieved from YouTube, and we retrieved these movie clips for research use only.

Figure~\ref{fig:Music2Video_2C} shows the selected results of music-to-video style transfer. We observe that different music genres transfer the clips into different color tones. Piano solo and symphony transfer the videos into a brighter tone, while chamber music transfer the videos into a darker tone. The genre of progressive house music is obviously not seen in the training set, so the transferred color tone is less usual compared to other samples. 
A common issue in this task is the fluctuation in brightness and color, which implies that the model is still unstable with the change of music features. To overcome this issue, additional constraints are needed to smooth the style in time domain, and this will be part of our future work.

\begin{figure}[t]
    \centering
    \includegraphics[width=\linewidth]{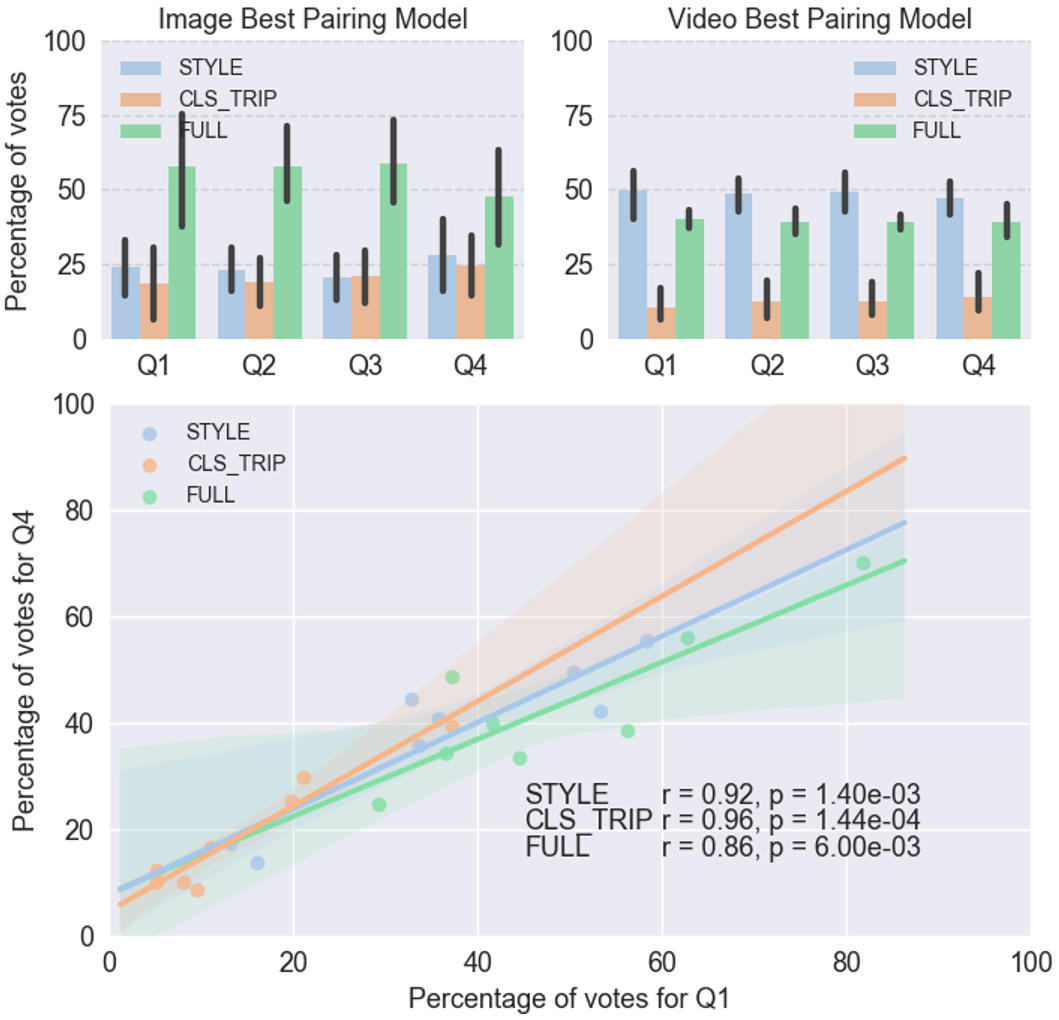}
    \caption{User study result. Upper left: best pairing model (in \%) on M2I samples. Upper right: best pairing model (in \%) on M2V samples. Bottom: correlation between Q1 and Q4.}
    \label{fig:SubEval}
\end{figure}

\vspace{-0.2cm}
\subsection{Aesthetic quality assessment: a user study}

To examine the visual aesthetics of the transferred images and videos, we refer to prior research on measuring people’s perceived aesthetic impressions  \cite{Moshagen2010visAes,Augustin2011visart,augustin2012all}.  
Key scholars summarize various dimensions of aesthetic judgements, from a general dimension “pleasingness” to commonly emphasized cognitive (e.g., comprehensibility, originality) and emotional (e.g., emotiveness, impressiveness) dimensions. We adopted questions from the emotional and pleasingness dimensions as the first step to assess visual aesthetics of the images (including painting, photograph) and videos after the transfer. We decided to focus on general perceived aesthetic impression instead of more technical aspects of visual aesthetics (e.g., saturation, contrast, stroke), as the latter type of assessment requires knowledge at the expert level.

We recruited a total of 137 participants (47.20\% male, 52.10\% female) to fill out an online questionnaire to assess eight image/video samples. Four are music-to-image (M2I) while four are music-to-video (M2V) samples. First, we had participants get familiar with the original images and audio/video content. Then, participants were asked to compare the style transfer outputs generated from three variants of our model: 1) style loss only (STYLE), 2) style loss plus the triplet and classification loss (CLS\_TRIP), and 3) the three loss terms plus the adversarial loss (FULL). It should be noted that we compared the original images/videos and STYLE in a pilot study but found that the clean images/videos could bias human judgement on the artifacts of the style transfer results. Therefore, we removed the images/ videos in the official study to reduce survey completion difficulty. For each image/ audio/ video content, participants were asked to select the one they perceived the most beautiful (Q1), attractive (Q2), moving (Q3), and most harmonious with the background music (Q4) out of the three images/videos style transfer outputs. Q1 and Q2 belong to the pleasingness dimension; Q3 belongs to the emotion dimension; Q4 is the objective of our research.

Results are shown in Figure~\ref{fig:SubEval}. For M2I samples, participants preferred the FULL model the most while for M2V samples, participants preferred the STYLE model. The difference in M2I and M2V reveals the roles of the objective functions in the generation process. Using only style loss in the M2I case seems to generate ‘boring’ result, but in the M2V case this is preferred as adding additional loss terms usually results in fluctuations in video. Comparing FULL to CLS\_TRIP, it shows that adding the adversarial loss helps both M2I and M2V. Additionally, participant responses to the four questions strongly correlated to each other. For instance, the correlation between the percentage of votes of Q1 and Q4 suggests that the model making the background music more harmonious with the visual style tends to make such visual style more beautiful (r = 0.92, 0.96 and 0.86 for the three models, respectively).

\section{Conclusion}%ding Remarks and Future Works}

We have demonstrated the feasibility of transferring visual styles directly from music. The flexibility of using different style transfer networks and of using either image or video contents in our framework all suggest great potential in the applications of animation and interactive arts. Evaluation results indicate the importance of a shared semantic space in solving the cross-modal style transfer problem, and also reveal the multi-dimensional nature of aesthetic quality assessment, which is still a challenging problem worth further study.

We have emphasized the importance of cross-model transfer in human creativity process. However, it should be noted that our proposed solution does not include all the scenarios that human artists deal with this problem in the real world. In fact, shared labels are not a necessary condition in real-world cross-model style transfer process. The condition of pairing an image to a music piece can usually be arbitrary and relies on how one interprets it. Shared labels are neither a sufficient condition in real-world cross-model style transfer. The reason that we can imagine a visual scene when listening to music is that our brains have built a complicated web of meaning that connects these data in various semantic levels, not because we know the time they were composed or painted. Annotations of high-level semantics such as art movement or genre could partly address this issue, but such annotations might be more difficult to be shared. On the other hand, the era labels adopted in this work ignore the
%Another concern which we have not consider in this work is 
the time asynchronization %and dependency of 
in the development of art and music (e.g., impressionism music appeared later than impressionism art). %We believe the artistic development of different domains is not always synchronous. Also, some 
%artistic concept and expression might influence across different decades. 
The purpose of this work is not to claim that learning the semantic links from era labels is the unique and `correct’ way to assign a style to an artwork. Rather, we emphasize that learning the arbitrary semantic links between different domains can be a feasible and scalable way for content generation. A future direction toward a more advanced cross-modal style transfer is to establish more label information such as genres and emotions to link the data from different modalities altogether up to a higher semantic level. 

\section*{Acknowledgement}
This work is partly supported by the Automatic Music Concert Animation (AMCA) project funded by the Institute of Information Science, Academia Sinica. 
We appreciate the hardware support by Medical Image Laboratory, National Chiao Tung University, and the support of music concert performance by KoKo Lab. 
We also thank those who participated in the subjective test and provided useful suggestions. 

\bibliographystyle{ACM-Reference-Format}
\balance
\bibliography{main}


\begin{thebibliography}{38}

%%% the Web address.

\ifx \showCODEN    \undefined \def \showCODEN     #1{\unskip}     \fi
\ifx \showDOI      \undefined \def \showDOI       #1{#1}\fi
\ifx \showISBNx    \undefined \def \showISBNx     #1{\unskip}     \fi
\ifx \showISBNxiii \undefined \def \showISBNxiii  #1{\unskip}     \fi
\ifx \showISSN     \undefined \def \showISSN      #1{\unskip}     \fi
\ifx \showLCCN     \undefined \def \showLCCN      #1{\unskip}     \fi
\ifx \shownote     \undefined \def \shownote      #1{#1}          \fi
\ifx \showarticletitle \undefined \def \showarticletitle #1{#1}   \fi
\ifx \showURL      \undefined \def \showURL       {\relax}        \fi
% invisible to TeX
\providecommand\bibfield[2]{#2}
\providecommand\bibinfo[2]{#2}
\providecommand\natexlab[1]{#1}
\providecommand\showeprint[2][]{arXiv:#2}

\bibitem[\protect\citeauthoryear{Augustin, Carbon, and Wagemans}{Augustin
  et~al\mbox{.}}{2011}]%
        {Augustin2011visart}
\bibfield{author}{\bibinfo{person}{M Augustin},
  \bibinfo{person}{Claus-Christian Carbon}, {and} \bibinfo{person}{Johan
  Wagemans}.} \bibinfo{year}{2011}\natexlab{}.
\newblock \showarticletitle{Measuring aesthetic impressions of visual art}.
\newblock \bibinfo{journal}{\emph{PERCEPTION}}  \bibinfo{volume}{40}
  (\bibinfo{date}{01} \bibinfo{year}{2011}), \bibinfo{pages}{219}.
\newblock


\bibitem[\protect\citeauthoryear{Augustin, Wagemans, and Carbon}{Augustin
  et~al\mbox{.}}{2012}]%
        {augustin2012all}
\bibfield{author}{\bibinfo{person}{M~Dorothee Augustin}, \bibinfo{person}{Johan
  Wagemans}, {and} \bibinfo{person}{Claus-Christian Carbon}.}
  \bibinfo{year}{2012}\natexlab{}.
\newblock \showarticletitle{All is beautiful? Generality vs. specificity of
  word usage in visual aesthetics}.
\newblock \bibinfo{journal}{\emph{Acta Psychologica}} \bibinfo{volume}{139},
  \bibinfo{number}{1} (\bibinfo{year}{2012}), \bibinfo{pages}{187--201}.
\newblock


\bibitem[\protect\citeauthoryear{Bergstrom, Karahalios, and Hart}{Bergstrom
  et~al\mbox{.}}{2007}]%
        {bergstrom2007isochords}
\bibfield{author}{\bibinfo{person}{Tony Bergstrom}, \bibinfo{person}{Karrie
  Karahalios}, {and} \bibinfo{person}{John~C Hart}.}
  \bibinfo{year}{2007}\natexlab{}.
\newblock \showarticletitle{Isochords: visualizing structure in music}. In
  \bibinfo{booktitle}{\emph{ACM Proc. Graphics Interface}}.
  \bibinfo{pages}{297--304}.
\newblock


\bibitem[\protect\citeauthoryear{Berman}{Berman}{1999}]%
        {berman1999synesthesia}
\bibfield{author}{\bibinfo{person}{Greta Berman}.}
  \bibinfo{year}{1999}\natexlab{}.
\newblock \showarticletitle{Synesthesia and the Arts}.
\newblock \bibinfo{journal}{\emph{Leonardo}} \bibinfo{volume}{32},
  \bibinfo{number}{1} (\bibinfo{year}{1999}), \bibinfo{pages}{15--22}.
\newblock


\bibitem[\protect\citeauthoryear{Brunner, Wang, Wattenhofer, and Zhao}{Brunner
  et~al\mbox{.}}{2018}]%
        {brunner2018symbolic}
\bibfield{author}{\bibinfo{person}{Gino Brunner}, \bibinfo{person}{Yuyi Wang},
  \bibinfo{person}{Roger Wattenhofer}, {and} \bibinfo{person}{Sumu Zhao}.}
  \bibinfo{year}{2018}\natexlab{}.
\newblock \showarticletitle{Symbolic music genre transfer with cyclegan}. In
  \bibinfo{booktitle}{\emph{ICTAI}}. \bibinfo{pages}{786--793}.
\newblock


\bibitem[\protect\citeauthoryear{Chelaramani, Jha, and Namboodiri}{Chelaramani
  et~al\mbox{.}}{2018}]%
        {chelaramani2018cross}
\bibfield{author}{\bibinfo{person}{Sahil Chelaramani},
  \bibinfo{person}{Abhishek Jha}, {and} \bibinfo{person}{Anoop~M. Namboodiri}.}
  \bibinfo{year}{2018}\natexlab{}.
\newblock \showarticletitle{Cross-Modal Style Transfer}. In
  \bibinfo{booktitle}{\emph{ICIP}}. \bibinfo{pages}{2157--2161}.
\newblock


\bibitem[\protect\citeauthoryear{Chen, Liao, Yuan, Yu, and Hua}{Chen
  et~al\mbox{.}}{2017}]%
        {chen2017coherent}
\bibfield{author}{\bibinfo{person}{Dongdong Chen}, \bibinfo{person}{Jing Liao},
  \bibinfo{person}{Lu Yuan}, \bibinfo{person}{Nenghai Yu}, {and}
  \bibinfo{person}{Gang Hua}.} \bibinfo{year}{2017}\natexlab{}.
\newblock \showarticletitle{Coherent online video style transfer}. In
  \bibinfo{booktitle}{\emph{ICCV}}. \bibinfo{pages}{1105--1114}.
\newblock


\bibitem[\protect\citeauthoryear{Collopy}{Collopy}{2000}]%
        {collopy2000color}
\bibfield{author}{\bibinfo{person}{Fred Collopy}.}
  \bibinfo{year}{2000}\natexlab{}.
\newblock \showarticletitle{Color, form, and motion: Dimensions of a musical
  art of light}.
\newblock \bibinfo{journal}{\emph{Leonardo}} \bibinfo{volume}{33},
  \bibinfo{number}{5} (\bibinfo{year}{2000}), \bibinfo{pages}{355--360}.
\newblock


\bibitem[\protect\citeauthoryear{Engel, Resnick, Roberts, Dieleman, Norouzi,
  Eck, and Simonyan}{Engel et~al\mbox{.}}{2017}]%
        {engel2017neural}
\bibfield{author}{\bibinfo{person}{Jesse Engel}, \bibinfo{person}{Cinjon
  Resnick}, \bibinfo{person}{Adam Roberts}, \bibinfo{person}{Sander Dieleman},
  \bibinfo{person}{Mohammad Norouzi}, \bibinfo{person}{Douglas Eck}, {and}
  \bibinfo{person}{Karen Simonyan}.} \bibinfo{year}{2017}\natexlab{}.
\newblock \showarticletitle{Neural audio synthesis of musical notes with
  wavenet autoencoders}. In \bibinfo{booktitle}{\emph{ICML}}.
  \bibinfo{pages}{1068--1077}.
\newblock


\bibitem[\protect\citeauthoryear{Gao, Gu, Zhang, and Yu}{Gao
  et~al\mbox{.}}{2018}]%
        {gao2018reconet}
\bibfield{author}{\bibinfo{person}{Chang Gao}, \bibinfo{person}{Derun Gu},
  \bibinfo{person}{Fangjun Zhang}, {and} \bibinfo{person}{Yizhou Yu}.}
  \bibinfo{year}{2018}\natexlab{}.
\newblock \showarticletitle{ReCoNet: Real-time Coherent Video Style Transfer
  Network}. In \bibinfo{booktitle}{\emph{ACCV}}. \bibinfo{pages}{637--653}.
\newblock


\bibitem[\protect\citeauthoryear{Gatys, Ecker, and Bethge}{Gatys
  et~al\mbox{.}}{2016}]%
        {gatys2016image}
\bibfield{author}{\bibinfo{person}{Leon~A Gatys}, \bibinfo{person}{Alexander~S
  Ecker}, {and} \bibinfo{person}{Matthias Bethge}.}
  \bibinfo{year}{2016}\natexlab{}.
\newblock \showarticletitle{Image style transfer using convolutional neural
  networks}. In \bibinfo{booktitle}{\emph{CVPR}}. \bibinfo{pages}{2414--2423}.
\newblock


\bibitem[\protect\citeauthoryear{Hawthorne, Huang, Ippolito, and Eck}{Hawthorne
  et~al\mbox{.}}{2018}]%
        {hawthorne2018transformer}
\bibfield{author}{\bibinfo{person}{Curtis Hawthorne}, \bibinfo{person}{Anna
  Huang}, \bibinfo{person}{Daphne Ippolito}, {and} \bibinfo{person}{Douglas
  Eck}.} \bibinfo{year}{2018}\natexlab{}.
\newblock \showarticletitle{Transformer-NADE for Piano Performances}. In
  \bibinfo{booktitle}{\emph{NIPS 2nd Workshop on Machine Learning for
  Creativity and Design}}.
\newblock


\bibitem[\protect\citeauthoryear{Huang and Belongie}{Huang and
  Belongie}{2017}]%
        {huang2017arbitrary}
\bibfield{author}{\bibinfo{person}{Xun Huang} {and} \bibinfo{person}{Serge
  Belongie}.} \bibinfo{year}{2017}\natexlab{}.
\newblock \showarticletitle{Arbitrary style transfer in real-time with adaptive
  instance normalization}. In \bibinfo{booktitle}{\emph{ICCV}}.
  \bibinfo{pages}{1501--1510}.
\newblock


\bibitem[\protect\citeauthoryear{Jolicoeur-Martineau}{Jolicoeur-Martineau}{2018}]%
        {jolicoeur2018relativistic}
\bibfield{author}{\bibinfo{person}{Alexia Jolicoeur-Martineau}.}
  \bibinfo{year}{2018}\natexlab{}.
\newblock \showarticletitle{The relativistic discriminator: a key element
  missing from standard GAN}.
\newblock \bibinfo{journal}{\emph{arXiv preprint arXiv:1807.00734}}
  (\bibinfo{year}{2018}).
\newblock


\bibitem[\protect\citeauthoryear{Kaliakatsos-Papakostas, Queiroz, Tsougras, and
  Cambouropoulos}{Kaliakatsos-Papakostas et~al\mbox{.}}{2017}]%
        {kaliakatsos2017conceptual}
\bibfield{author}{\bibinfo{person}{Maximos Kaliakatsos-Papakostas},
  \bibinfo{person}{Marcelo Queiroz}, \bibinfo{person}{Costas Tsougras}, {and}
  \bibinfo{person}{Emilios Cambouropoulos}.} \bibinfo{year}{2017}\natexlab{}.
\newblock \showarticletitle{Conceptual blending of harmonic spaces for creative
  melodic harmonisation}.
\newblock \bibinfo{journal}{\emph{JNMR}} \bibinfo{volume}{46},
  \bibinfo{number}{4} (\bibinfo{year}{2017}), \bibinfo{pages}{305--328}.
\newblock


\bibitem[\protect\citeauthoryear{Kennedy}{Kennedy}{2007}]%
        {kennedy2007painting}
\bibfield{author}{\bibinfo{person}{Sharon~L Kennedy}.}
  \bibinfo{year}{2007}\natexlab{}.
\newblock \showarticletitle{Painting music: rhythm and movement in art}.
\newblock  (\bibinfo{year}{2007}).
\newblock


\bibitem[\protect\citeauthoryear{Kobayashi, Toda, Neubig, Sakti, and
  Nakamura}{Kobayashi et~al\mbox{.}}{2014}]%
        {kobayashi2014statistical}
\bibfield{author}{\bibinfo{person}{Kazuhiro Kobayashi}, \bibinfo{person}{Tomoki
  Toda}, \bibinfo{person}{Graham Neubig}, \bibinfo{person}{Sakriani Sakti},
  {and} \bibinfo{person}{Satoshi Nakamura}.} \bibinfo{year}{2014}\natexlab{}.
\newblock \showarticletitle{Statistical singing voice conversion with direct
  waveform modification based on the spectrum differential}. In
  \bibinfo{booktitle}{\emph{ISCA}}.
\newblock


\bibitem[\protect\citeauthoryear{Li, Liu, Kautz, and Yang}{Li
  et~al\mbox{.}}{2018}]%
        {li2018learning}
\bibfield{author}{\bibinfo{person}{Xueting Li}, \bibinfo{person}{Sifei Liu},
  \bibinfo{person}{Jan Kautz}, {and} \bibinfo{person}{Ming-Hsuan Yang}.}
  \bibinfo{year}{2018}\natexlab{}.
\newblock \showarticletitle{Learning linear transformations for fast arbitrary
  style transfer}.
\newblock \bibinfo{journal}{\emph{arXiv:1808.04537}} (\bibinfo{year}{2018}).
\newblock


\bibitem[\protect\citeauthoryear{Li, Fang, Yang, Wang, Lu, and Yang}{Li
  et~al\mbox{.}}{2017}]%
        {li2017universal}
\bibfield{author}{\bibinfo{person}{Yijun Li}, \bibinfo{person}{Chen Fang},
  \bibinfo{person}{Jimei Yang}, \bibinfo{person}{Zhaowen Wang},
  \bibinfo{person}{Xin Lu}, {and} \bibinfo{person}{Ming-Hsuan Yang}.}
  \bibinfo{year}{2017}\natexlab{}.
\newblock \showarticletitle{Universal style transfer via feature transforms}.
  In \bibinfo{booktitle}{\emph{NeurIPS}}. \bibinfo{pages}{386--396}.
\newblock


\bibitem[\protect\citeauthoryear{Lu, Xue, Chang, Lee, and Su}{Lu
  et~al\mbox{.}}{2018}]%
        {lu2018play}
\bibfield{author}{\bibinfo{person}{Chien-Yu Lu}, \bibinfo{person}{Min-Xin Xue},
  \bibinfo{person}{Chia-Che Chang}, \bibinfo{person}{Che-Rung Lee}, {and}
  \bibinfo{person}{Li Su}.} \bibinfo{year}{2018}\natexlab{}.
\newblock \showarticletitle{Play as You Like: Timbre-enhanced Multi-modal Music
  Style Transfer}.
\newblock \bibinfo{journal}{\emph{arXiv preprint arXiv:1811.12214}}
  (\bibinfo{year}{2018}).
\newblock


\bibitem[\protect\citeauthoryear{Lu and Su}{Lu and Su}{2018}]%
        {lu2018transferring}
\bibfield{author}{\bibinfo{person}{Wei~Tsung Lu} {and} \bibinfo{person}{Li
  Su}.} \bibinfo{year}{2018}\natexlab{}.
\newblock \showarticletitle{Transferring the Style of Homophonic Music Using
  Recurrent Neural Networks and Autoregressive Model.}. In
  \bibinfo{booktitle}{\emph{ISMIR}}. \bibinfo{pages}{740--746}.
\newblock


\bibitem[\protect\citeauthoryear{Maezawa}{Maezawa}{2018}]%
        {maezawa2018deep}
\bibfield{author}{\bibinfo{person}{A Maezawa}.}
  \bibinfo{year}{2018}\natexlab{}.
\newblock \showarticletitle{Deep piano performance rendering with conditional
  VAE}. In \bibinfo{booktitle}{\emph{ISMIR Late Breaking and Demo Papers}}.
\newblock


\bibitem[\protect\citeauthoryear{Malik and Ek}{Malik and Ek}{2017}]%
        {malik2017neural}
\bibfield{author}{\bibinfo{person}{Iman Malik} {and}
  \bibinfo{person}{Carl~Henrik Ek}.} \bibinfo{year}{2017}\natexlab{}.
\newblock \showarticletitle{Neural translation of musical style}.
\newblock \bibinfo{journal}{\emph{arXiv preprint arXiv:1708.03535}}
  (\bibinfo{year}{2017}).
\newblock


\bibitem[\protect\citeauthoryear{Mardirossian and Chew}{Mardirossian and
  Chew}{2007}]%
        {mardirossian2007visualizing}
\bibfield{author}{\bibinfo{person}{Arpi Mardirossian} {and}
  \bibinfo{person}{Elaine Chew}.} \bibinfo{year}{2007}\natexlab{}.
\newblock \showarticletitle{Visualizing Music: Tonal Progressions and
  Distributions.}. In \bibinfo{booktitle}{\emph{ISMIR}}. Citeseer,
  \bibinfo{pages}{189--194}.
\newblock


\bibitem[\protect\citeauthoryear{Moshagen and Thielsch}{Moshagen and
  Thielsch}{2010}]%
        {Moshagen2010visAes}
\bibfield{author}{\bibinfo{person}{Morten Moshagen} {and}
  \bibinfo{person}{Meinald~T. Thielsch}.} \bibinfo{year}{2010}\natexlab{}.
\newblock \showarticletitle{Facets of Visual Aesthetics}.
\newblock \bibinfo{journal}{\emph{Int. J. Hum.-Comput. Stud.}}
  \bibinfo{volume}{68}, \bibinfo{number}{10} (\bibinfo{date}{Oct.}
  \bibinfo{year}{2010}), \bibinfo{pages}{689–709}.
\newblock
\showISSN{1071-5819}
\urldef\tempurl%
\url{https://doi.org/10.1016/j.ijhcs.2010.05.006}
\showDOI{\tempurl}


\bibitem[\protect\citeauthoryear{Peacock}{Peacock}{1988}]%
        {peacock1988instruments}
\bibfield{author}{\bibinfo{person}{Kenneth Peacock}.}
  \bibinfo{year}{1988}\natexlab{}.
\newblock \showarticletitle{Instruments to perform color-music: Two centuries
  of technological experimentation}.
\newblock \bibinfo{journal}{\emph{Leonardo}} \bibinfo{volume}{21},
  \bibinfo{number}{4} (\bibinfo{year}{1988}), \bibinfo{pages}{397--406}.
\newblock


\bibitem[\protect\citeauthoryear{Ronneberger, Fischer, and Brox}{Ronneberger
  et~al\mbox{.}}{2015}]%
        {ronneberger2015u}
\bibfield{author}{\bibinfo{person}{Olaf Ronneberger}, \bibinfo{person}{Philipp
  Fischer}, {and} \bibinfo{person}{Thomas Brox}.}
  \bibinfo{year}{2015}\natexlab{}.
\newblock \showarticletitle{U-net: Convolutional networks for biomedical image
  segmentation}. In \bibinfo{booktitle}{\emph{International Conference on
  Medical image computing and computer-assisted intervention}}. Springer,
  \bibinfo{pages}{234--241}.
\newblock


\bibitem[\protect\citeauthoryear{{Schroff}, {Kalenichenko}, and
  {Philbin}}{{Schroff} et~al\mbox{.}}{2015}]%
        {schroff2015facenet}
\bibfield{author}{\bibinfo{person}{F. {Schroff}}, \bibinfo{person}{D.
  {Kalenichenko}}, {and} \bibinfo{person}{J. {Philbin}}.}
  \bibinfo{year}{2015}\natexlab{}.
\newblock \showarticletitle{FaceNet: A unified embedding for face recognition
  and clustering}. In \bibinfo{booktitle}{\emph{CVPR}}.
  \bibinfo{pages}{815--823}.
\newblock
\showISSN{1063-6919}
\urldef\tempurl%
\url{https://doi.org/10.1109/CVPR.2015.7298682}
\showDOI{\tempurl}


\bibitem[\protect\citeauthoryear{Shen, Lei, Barzilay, and Jaakkola}{Shen
  et~al\mbox{.}}{2017}]%
        {shen2017style}
\bibfield{author}{\bibinfo{person}{Tianxiao Shen}, \bibinfo{person}{Tao Lei},
  \bibinfo{person}{Regina Barzilay}, {and} \bibinfo{person}{Tommi Jaakkola}.}
  \bibinfo{year}{2017}\natexlab{}.
\newblock \showarticletitle{Style transfer from non-parallel text by
  cross-alignment}. In \bibinfo{booktitle}{\emph{NeurIPS}}.
  \bibinfo{pages}{6830--6841}.
\newblock


\bibitem[\protect\citeauthoryear{Sheng, Lin, Shao, and Wang}{Sheng
  et~al\mbox{.}}{2018}]%
        {sheng2018avatar}
\bibfield{author}{\bibinfo{person}{Lu Sheng}, \bibinfo{person}{Ziyi Lin},
  \bibinfo{person}{Jing Shao}, {and} \bibinfo{person}{Xiaogang Wang}.}
  \bibinfo{year}{2018}\natexlab{}.
\newblock \showarticletitle{Avatar-net: Multi-scale zero-shot style transfer by
  feature decoration}. In \bibinfo{booktitle}{\emph{CVPR}}.
  \bibinfo{pages}{8242--8250}.
\newblock


\bibitem[\protect\citeauthoryear{Simonyan and Zisserman}{Simonyan and
  Zisserman}{2014}]%
        {simonyan2014vgg}
\bibfield{author}{\bibinfo{person}{Karen Simonyan} {and}
  \bibinfo{person}{Andrew Zisserman}.} \bibinfo{year}{2014}\natexlab{}.
\newblock \showarticletitle{Very Deep Convolutional Networks for Large-Scale
  Image Recognition}.
\newblock \bibinfo{journal}{\emph{arXiv 1409.1556}} (\bibinfo{date}{09}
  \bibinfo{year}{2014}).
\newblock


\bibitem[\protect\citeauthoryear{Verma and Smith}{Verma and Smith}{2018}]%
        {verma2018neural}
\bibfield{author}{\bibinfo{person}{Prateek Verma} {and}
  \bibinfo{person}{Julius~O Smith}.} \bibinfo{year}{2018}\natexlab{}.
\newblock \showarticletitle{Neural style transfer for audio spectograms}.
\newblock \bibinfo{journal}{\emph{arXiv preprint arXiv:1801.01589}}
  (\bibinfo{year}{2018}).
\newblock


\bibitem[\protect\citeauthoryear{Wan, Chuang, and Lee}{Wan
  et~al\mbox{.}}{2019}]%
        {wan2019towards}
\bibfield{author}{\bibinfo{person}{Chia-Hung Wan}, \bibinfo{person}{Shun-Po
  Chuang}, {and} \bibinfo{person}{Hung-Yi Lee}.}
  \bibinfo{year}{2019}\natexlab{}.
\newblock \showarticletitle{Towards Audio to Scene Image Synthesis Using
  Generative Adversarial Network}. In \bibinfo{booktitle}{\emph{ICASSP}}.
  \bibinfo{pages}{496--500}.
\newblock


\bibitem[\protect\citeauthoryear{Wu, Liu, Yang, and Jang}{Wu
  et~al\mbox{.}}{2018}]%
        {wu2018singing}
\bibfield{author}{\bibinfo{person}{Cheng-Wei Wu}, \bibinfo{person}{Jen-Yu Liu},
  \bibinfo{person}{Yi-Hsuan Yang}, {and} \bibinfo{person}{Jyh-Shing~R Jang}.}
  \bibinfo{year}{2018}\natexlab{}.
\newblock \showarticletitle{Singing Style Transfer Using Cycle-Consistent
  Boundary Equilibrium Generative Adversarial Networks}.
\newblock \bibinfo{journal}{\emph{arXiv preprint arXiv:1807.02254}}
  (\bibinfo{year}{2018}).
\newblock


\bibitem[\protect\citeauthoryear{Zhang, Goodfellow, Metaxas, and Odena}{Zhang
  et~al\mbox{.}}{2018}]%
        {zhang2018self}
\bibfield{author}{\bibinfo{person}{Han Zhang}, \bibinfo{person}{Ian
  Goodfellow}, \bibinfo{person}{Dimitris Metaxas}, {and}
  \bibinfo{person}{Augustus Odena}.} \bibinfo{year}{2018}\natexlab{}.
\newblock \showarticletitle{Self-attention generative adversarial networks}.
\newblock \bibinfo{journal}{\emph{arXiv preprint arXiv:1805.08318}}
  (\bibinfo{year}{2018}).
\newblock


\bibitem[\protect\citeauthoryear{Zhao, Gan, Rouditchenko, Vondrick, McDermott,
  and Torralba}{Zhao et~al\mbox{.}}{2018}]%
        {zhao2018sound}
\bibfield{author}{\bibinfo{person}{Hang Zhao}, \bibinfo{person}{Chuang Gan},
  \bibinfo{person}{Andrew Rouditchenko}, \bibinfo{person}{Carl Vondrick},
  \bibinfo{person}{Josh McDermott}, {and} \bibinfo{person}{Antonio Torralba}.}
  \bibinfo{year}{2018}\natexlab{}.
\newblock \showarticletitle{The sound of pixels}. In
  \bibinfo{booktitle}{\emph{ECCV}}. \bibinfo{pages}{570--586}.
\newblock


\bibitem[\protect\citeauthoryear{Zhou, Wang, Fang, Bui, and Berg}{Zhou
  et~al\mbox{.}}{2018}]%
        {zhou2018visual}
\bibfield{author}{\bibinfo{person}{Yipin Zhou}, \bibinfo{person}{Zhaowen Wang},
  \bibinfo{person}{Chen Fang}, \bibinfo{person}{Trung Bui}, {and}
  \bibinfo{person}{Tamara~L Berg}.} \bibinfo{year}{2018}\natexlab{}.
\newblock \showarticletitle{Visual to sound: Generating natural sound for
  videos in the wild}. In \bibinfo{booktitle}{\emph{CVPR}}.
  \bibinfo{pages}{3550--3558}.
\newblock


\bibitem[\protect\citeauthoryear{Zhu, Park, Isola, and Efros}{Zhu
  et~al\mbox{.}}{2017}]%
        {zhu2017unpaired}
\bibfield{author}{\bibinfo{person}{Jun-Yan Zhu}, \bibinfo{person}{Taesung
  Park}, \bibinfo{person}{Phillip Isola}, {and} \bibinfo{person}{Alexei~A
  Efros}.} \bibinfo{year}{2017}\natexlab{}.
\newblock \showarticletitle{Unpaired image-to-image translation using
  cycle-consistent adversarial networks}. In
  \bibinfo{booktitle}{\emph{Proceedings of the IEEE international conference on
  computer vision}}. \bibinfo{pages}{2223--2232}.
\newblock


\end{thebibliography}

\end{document}